\documentstyle[jair,twoside,11pt,theapa,url]{article}

\input epsf %% at top of file

\newcommand{\psfigxx}[6]{
        \begin{figure}[#5]
        \begin{center}
        \epsfxsize = #6 in
        \leavevmode\epsfbox{#2}
        \end{center}
        \vspace{-#3in}
        \caption{#4}
        \label{#1}
        \end{figure}
}

\newcommand{\psfigyy}[6]{
        \begin{figure}[#5]
        \begin{center}
        \epsfxsize = #6 in
        \leavevmode\epsfbox{#2}
        \end{center}
        \vspace{#3in}
        \caption{#4}
        \label{#1}
        \end{figure}
}

\jairheading{29}{2007}{105-151}{05/06}{06/07}
\ShortHeadings{Combination Strategies for Semantic Role Labeling}
{Surdeanu, M{\`a}rquez, Carreras, \& Comas}
\firstpageno{105}

\begin{document}

\title{Combination Strategies for Semantic Role Labeling}

\author{\name Mihai Surdeanu \email surdeanu@lsi.upc.edu \\
       \name Llu{\'{\i}}s M{\`a}rquez \email lluism@lsi.upc.edu \\
       \name Xavier Carreras \email carreras@lsi.upc.edu \\
       \name Pere R. Comas \email pcomas@lsi.upc.edu \\
       \addr Technical University of Catalonia,\\
        C/ Jordi Girona, 1-3\\
        08034 Barcelona, SPAIN}

% For research notes, remove the comment character in the line below.
% \researchnote

\maketitle

\begin{abstract}
  
  This paper introduces and analyzes a battery of inference models for
  the problem of semantic role labeling: one based on constraint
  satisfaction, and several strategies that model the inference as a
  meta-learning problem using discriminative classifiers. These
  classifiers are developed with a rich set of novel features that
  encode proposition and sentence-level information.
  To our knowledge, this is the first
  work that: (a) performs a thorough analysis of learning-based
  inference models for semantic role labeling, and (b) compares
  several inference strategies in this context.
  We evaluate the proposed inference strategies in the framework of
  the CoNLL-2005 shared task using only automatically-generated
  syntactic information.  The extensive experimental evaluation
  and analysis
  indicates that all the proposed inference strategies are successful
  $-$they all outperform the current best results reported in the CoNLL-2005
  evaluation exercise$-$
  but each of the proposed approaches has its advantages and disadvantages.
Several important traits of a state-of-the-art SRL combination strategy 
emerge from this analysis:
(i) individual models should be combined at the granularity of
  candidate arguments rather than at the granularity of complete solutions; 
(ii) the best combination strategy uses an inference model based in
  learning; and (iii) the learning-based inference benefits from
  max-margin classifiers and  global feedback. 

\end{abstract}

\section{Introduction}
\label{sec:introduction}

Natural Language Understanding (NLU) is a subfield of Artificial
Intelligence (AI) that deals with the extraction of the semantic
information available in natural language texts.
This knowledge is used to develop high-level applications requiring
textual and document understanding, such as Question Answering or
Information Extraction.
NLU is a complex
``AI-complete'' problem that needs to venture well beyond the
syntactic analysis of natural language texts. While the state of the
art in NLU is still far from reaching its goals, recent research has
made important progress in a subtask of NLU: Semantic Role Labeling.
The task of Semantic Role Labeling (SRL) is the process of
detecting basic event structures such as {\em who} did {\em what} to
{\em whom}, {\em when} and {\em where}. See Figure~\ref{fig:pb} for a
sample sentence annotated with such an event frame.

\subsection{Motivation}

SRL has received considerable interest in the past few
years~\cite{gil02,sur03,xue04,pra05,car05}. It was shown that
the identification of such event frames has a significant contribution
for many NLU applications such as
Information Extraction~\cite{sur03}, Question
Answering~\cite{nar04}, Machine Translation~\cite{boa02},
Summarization~\cite{mwl+05}, and Coreference Resolution~\cite{pon06a,pon06b}.

From a syntactic perspective, most machine-learning SRL approaches can be
classified in one of two classes: 
approaches that take advantage of complete syntactic analysis of text,
pioneered by~\citeA{gil02}, and approaches that use partial syntactic
analysis, championed by previous evaluations performed within the
Conference on Computational Natural Language Learning
(CoNLL)~(Carreras \& M{\`a}rquez, 2004, 2005).
%~\cite{car04,car05}. 
The wisdom extracted from the first representation
indicates that full syntactic analysis has a significant contribution
to SRL performance, {\em when using hand-corrected syntactic
information}~\cite{gil02b}.
On the other hand, when only automatically-generated syntax is
available, the quality of the information provided through full syntax
decreases because the state-of-the-art of full parsing is less robust
and performs worse than the tools used for partial syntactic
analysis. Under such real-world conditions, the difference between the
two SRL approaches (with full or partial syntax) is not that
high. 
More interestingly, {\em the two SRL strategies perform better for
different semantic roles}. For example, models that use full syntax
recognize agent and theme roles better, whereas models based on
partial syntax are better at recognizing explicit patient roles,
which tend to be farther from the predicate and accumulate more
parsing errors~\cite{mar05}.

\subsection{Approach}

In this article we explore the implications of the above observations
by studying strategies for combining the output of several independent
SRL systems, which take advantage of different syntactic views of the
text.  In a given sentence, our combination models receive labeled
arguments from individual systems, and produce an overall argument
structure for the corresponding sentence.  
The proposed combination strategies
exploit several levels of information: local and global features (from
individual models) and constraints on the argument structure.  In this
work, we investigate three different approaches:
\begin{itemize}
\vspace{-2mm}
\item The first combination model has no parameters to estimate; it
  only makes use of the argument probabilities 
output by the individual models and constraints
over argument structures to build the overall solution for each sentence.
We call this model {\em inference with constraint satisfaction}.
\vspace{-2mm}
\item The second approach implements a cascaded inference model with
local learning: first, for each type of argument, a classifier trained
offline decides
whether a candidate is or is not a final argument. Next, the
candidates that passed the previous step are combined into a
solution consistent with the constraints over argument structures.
We refer to this model as {\em inference with local learning}.
\vspace{-2mm}
\item The third inference model is global:
a number of online ranking functions, one for each argument type, are trained
to score argument candidates so that the correct argument structure 
for the complete sentence is
globally ranked at the top. We call this model {\em inference with
  global learning}.
\end{itemize}
The proposed combination strategies are general and do not depend on
the way in which candidate arguments are collected. We empirically
prove it by experimenting not only with individual SRL systems
developed in house, but also with the 10 best systems at the 
CoNLL-2005 shared task evaluation.

\psfigxx{fig:pb}{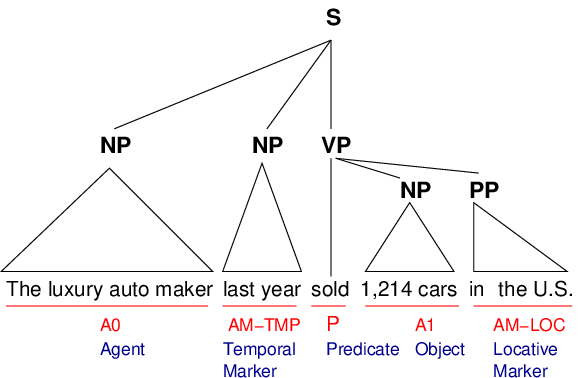}{0}{Sample sentence from the PropBank
  corpus.}{t}{4} 

\subsection{Contribution}

The work introduced in this paper has several novel points. To our
knowledge, this is the first work that thoroughly explores an
inference model based on meta-learning (the second and third inference
models introduced) in the context of SRL. We investigate meta-learning
combination strategies based on rich, global representations in the
form of local and global features, and in the form of structural
constraints of solutions.  Our empirical analysis indicates that these
combination strategies outperform the current state of the art.  Note
that all the combination strategies proposed in this paper are not
``re-ranking'' approaches~\cite{hag05,col00}. Whereas re-ranking selects
the overall best solution from a pool of {\em complete} solutions of
the individual models, our combination approaches combine candidate
arguments, or {\em incomplete solutions}, from {\em different}
individual models. We show that our approach has better potential,
i.e., the upper limit on the F$_1$ score is higher
and performance is better on several corpora.

A second novelty of this paper is that it performs a comparative
analysis of several combination strategies for SRL, using the same
framework $-$i.e., the same pool of candidates$-$ and the same evaluation
methodology. While a large number of combination approaches have been
previously analyzed in the context of SRL or in the larger context of
predicting structures in natural language texts $-$e.g., inference based on
constraint satisfaction~\cite{koo05,rot05}, inference based in local
learning~\cite{mar05}, re-ranking~\cite{col00,hag05} etc.$-$ it is
still not clear which strategy performs best for semantic role
labeling. In this paper we provide empirical answers to several
important questions in this respect. For example, is a combination
strategy based on constraint satisfaction better than an inference
model based on learning? Or, how important is global feedback in the
learning-based inference model?
Our analysis indicates that the following issues are important traits
of a state-of-the-art combination SRL system: 
(i) the individual models are combined at argument granularity rather
than at the granularity of complete solutions (typical of re-ranking);  
(ii) the best combination strategy uses an inference model based in learning;
and (iii) the learning-based inference benefits from max-margin
classifiers and global feedback. 

The paper is organized as follows. Section~\ref{sec:corpora}
introduces the semantic corpora used for training and evaluation.
Section~\ref{sec:approach} overviews the proposed combination
approaches. The individual SRL models are introduced in
Section~\ref{sec:base} and evaluated in Section~\ref{sec:indperf}.
Section~\ref{sec:features} lists the features used by the three
combination models introduced in this paper. The combination models
themselves are described in Section~\ref{sec:models}.
Section~\ref{sec:results} introduces an empirical analysis of the
proposed combination methods. Section~\ref{sec:related} reviews
related work and Section~\ref{sec:conclusions} concludes the paper.
%%
%% End intro
%%%%%%%%%%%%%%

\section{Semantic Corpora}
\label{sec:corpora}

In this paper we have used PropBank, an approximately
one-million-word corpus annotated with predicate-argument
structures~\cite{pal05}. To date, PropBank addresses only predicates
lexicalized by verbs.
Besides predicate-argument structures, PropBank contains full
syntactic analysis of its sentences, because it extends the Wall
Street Journal (WSJ) part of the Penn Treebank, a corpus that was
previously annotated with syntactic information~\cite{mar94}. 

For any given predicate, a survey was carried out to determine the
predicate usage, and, if required, the usages were divided into major
senses. However, the senses are divided more on syntactic grounds than
semantic, following the assumption that syntactic frames are a direct
reflection of underlying semantics.
The arguments of each predicate are numbered sequentially from {\tt
A0} to {\tt A5}. Generally, {\tt A0} stands for {\em agent},
{\tt A1} for {\em theme} or {\em direct object}, and {\tt A2} for
{\em indirect object}, {\em benefactive} or {\em instrument}, but
semantics tend to be verb specific. Additionally, predicates might
have adjunctive arguments, referred to as {\tt AM}s. For
example, {\tt AM-LOC} indicates a locative and {\tt AM-TMP}
indicates a temporal.  Figure~\ref{fig:pb} shows a sample
PropBank sentence where one predicate (``sold'') has 4 arguments.
Both regular and adjunctive arguments can be discontinuous, in which
case the trailing argument fragments are prefixed by {\tt C-},
e.g., ``[$_{\small{\tt A1}}$ Both funds] are [$_{\small{\tt
      predicate}}$ expected] [$_{\small{\tt C-A1}}$ to begin operation
  around March 1].'' Finally, PropBank contains argument references
(typically pronominal), which share the same label with the actual
argument prefixed with {\tt R-}.\footnote{In the original PropBank annotations, co-referenced arguments appear as a single item, with no differentiation between the referent and the reference. Here we use the version of the data used in the CoNLL shared tasks, where reference arguments were automatically separated
from their corresponding referents with simple pattern-matching rules.}

In this paper we do not use
any syntactic information from the Penn Treebank. Instead, we develop our
models using automatically-generated syntax and named-entity (NE)
labels, made available by the CoNLL-2005 shared task
evaluation~\cite{car05}. From the CoNLL data, we use the syntactic
trees generated by the Charniak parser~\cite{cha00} to develop two
individual models based on full syntactic analysis, and the chunk
$-$i.e., basic syntactic phrase$-$ labels and clause boundaries to
construct a partial-syntax model. All individual models use 
the provided NE labels.  

Switching from hand-corrected to automatically-generated syntactic
information means that the PropBank assumption that each
argument (or argument fragment for discontinuous arguments)
maps to one syntactic phrase no longer holds, due to errors of the
syntactic processors. Our analysis of the PropBank data indicates that
only 91.36\% of the semantic arguments can be matched to exactly one
phrase generated by the Charniak parser. Essentially, this means that
SRL approaches that make the assumption that each semantic argument
maps to one syntactic construct can not recognize almost 9\% of the
arguments. The same statement can be made about approaches based on partial
syntax with the caveat that in this setup arguments have to match a
sequence of chunks. However, one expects that the degree of
compatibility between syntactic chunks and semantic arguments is
higher due to the finer granularity of the syntactic elements and
because chunking algorithms perform better than full parsing 
algorithms. Indeed, our analysis of the same PropBank data supports
this observation: 95.67\% of the semantic arguments can be matched to
a sequence of chunks generated by the CoNLL syntactic chunker.  

Following the CoNLL-2005 setting we evaluated our system not only on
PropBank but also on a fresh test set, derived from the Brown
corpus. This second evaluation allows us to  
investigate the robustness of the proposed combination models.

\section{Overview of the Combination Strategies}
\label{sec:approach}

In this paper we introduce and analyze three combination strategies
for the problem of semantic role labeling. The three combination
strategies are implemented on a shared framework $-$detailed in
Figure~\ref{fig:arch}$-$ which consists of several stages: (a)
generation of candidate arguments, (b) candidate scoring, and finally
(c) inference. For clarity, we describe first the proposed combination
framework, i.e., the vertical flow in Figure~\ref{fig:arch}. Then,
we move to an overview of the three combination
methodologies, shown horizontally in Figure~\ref{fig:arch}. 

\psfigxx{fig:arch}{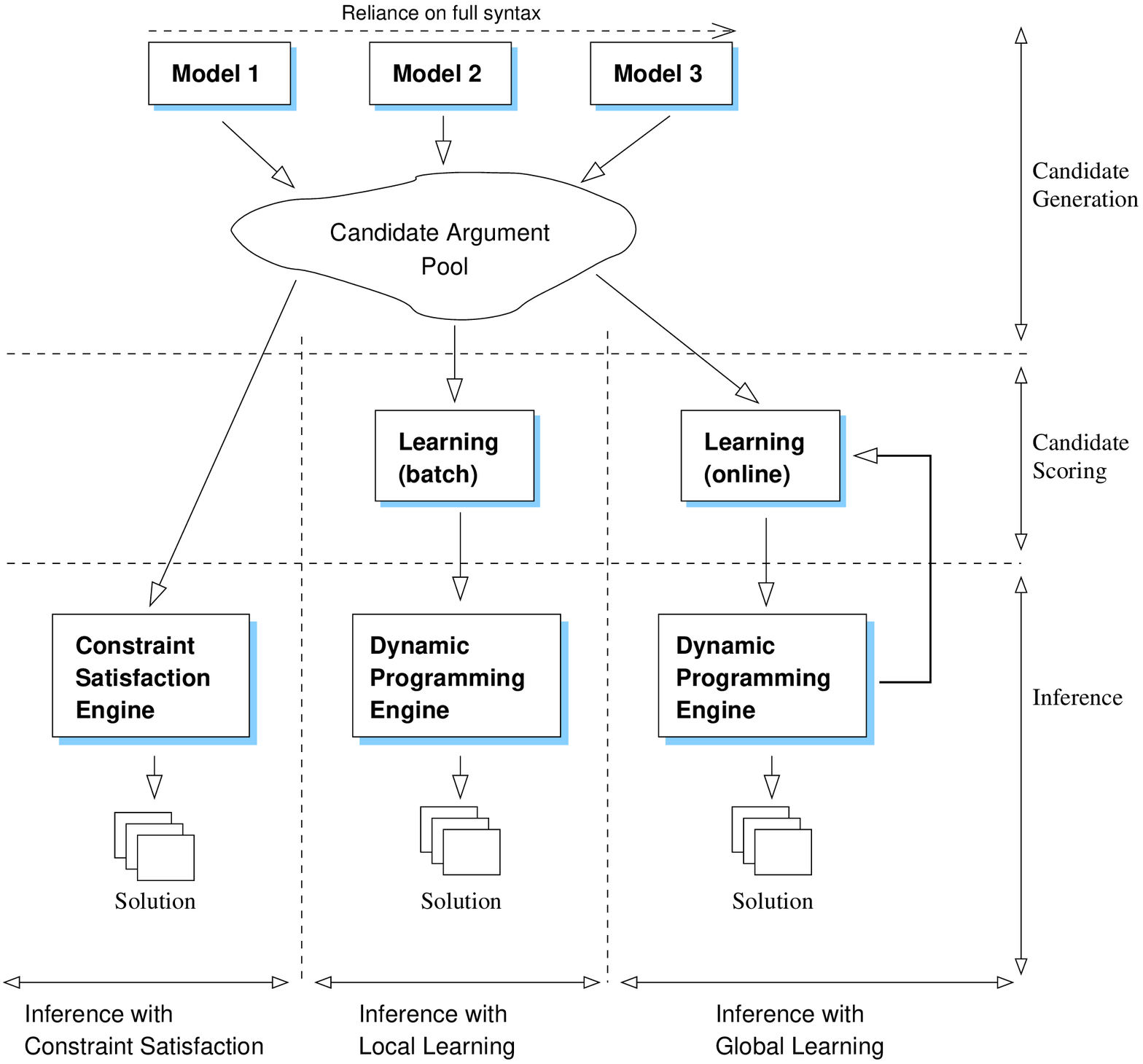}{0.15}{Overview of the proposed
  combination strategies.}{t!}{5.0} 

In the {\em candidate generation} step, we merge the solutions of three
individual SRL models into a unique pool of candidate arguments. The
individual SRL models range from complete reliance on full parsing to using
only partial syntactic information. For example, Model~1 is developed
as a sequential tagger (using the B-I-O tagging scheme) with only
partial syntactic information (basic phrases and clause boundaries),
whereas Model~3 uses full syntactic analysis of the text and handles
only arguments that map into exactly one syntactic
constituent. We detail the individual SRL models in
Section~\ref{sec:base} and empirically evaluate them in
Section~\ref{sec:indperf}. 

In the {\em candidate scoring} phrase, we re-score all
candidate arguments using both local information, e.g., the syntactic
structure of the candidate argument, and global information, e.g., how
many individual models have generated similar candidate arguments. We
describe all the features used for candidate scoring in
Section~\ref{sec:features}. 

Finally, in the {\em inference} stage the combination models search
for the best solution that 
is consistent with the domain constraints, e.g., two arguments for the
same predicate cannot overlap or embed, a predicate may not
have more than one core argument ({\tt A0-5}), etc. 

All the combination approaches proposed in this paper share the same 
candidate argument pool. This guarantees that the results obtained by
the different strategies on the same corpus are comparable. On the
other hand, even though the candidate generation step is shared, the
three combination methodologies differ significantly in their scoring
and inference models. 

The first combination strategy analyzed, {\em inference with
constraint satisfaction}, skips the candidate scoring step
completely and uses instead the probabilities output by the individual SRL
models for each candidate argument. If the individual models' raw
activations are not 
actual probabilities we convert them to probabilities using the
{\em softmax} function~\cite{bish95}, before passing them to the
inference component. The inference is implemented using a Constraint
Satisfaction model that searches for the solution that maximizes a
certain compatibility function. 
The compatibility function models not only the
probability of the global solution but also the consistency of the
solution according to the domain constraints. This combination
strategy is based on the technique presented by~\citeA{koo05}. 
The main difference between the two systems is in
the candidate generation step: we use three
independent individual SRL models, whereas Komen et al. used the same
SRL model trained on different syntactic views of the data, i.e., the
top parse trees generated by the Charniak and Collins parsers~\cite{cha00,col99b}. 
Furthermore, we take our argument candidates from the set of complete
solutions generated by the individual models, whereas Komen et al.
take them from different syntactic trees, before constructing any
complete solution.
The obvious advantage of the inference model with Constraint Satisfaction
is that it is unsupervised: no learning is necessary for candidate
scoring, because the scores of the individual models are used. 
On the other hand, the Constraint Satisfaction model requires
that the individual models provide raw activations, and, moreover,
that the raw activations be convertible to true probabilities. 

The second combination strategy proposed in this article, {\em
inference with local learning}, 
re-scores all candidates in the pool using a set of binary
discriminative classifiers. 
The classifiers assign to each argument a score measuring the 
confidence that the argument is part of the correct, global
solution. The classifiers are trained in batch mode
and are completely decoupled from the inference module. The inference
component is implemented using a CKY-based dynamic programming
algorithm~\cite{you67}. The main advantage of this strategy is that
candidates are re-scored using significantly more information than
what is available to each individual model. For example, we
incorporate features that count the number of individual systems that
generated the given candidate argument, several types of overlaps with
candidate arguments of the same predicate and also with arguments of
other predicates, 
structural information based on both full and partial syntax,
etc. We describe the rich feature set used for the scoring of
candidate arguments in Section~\ref{sec:features}. Also, this
combination approach does not depend on the argument probabilities of the
individual SRL models (but can incorporate them as features, if
available). 
This combination approach is more complex than the previous strategy 
because it has an additional step that requires
supervised learning: candidate scoring. Nevertheless, this
does not mean that additional corpus is necessary:
using cross validation,
the candidate scoring classifiers can be trained on the same corpus
used to train the individual SRL models. Moreover, we show in
Section~\ref{sec:results} that we obtain excellent performance even
when the candidate scoring classifiers are trained on significantly
less data than the individual SRL models. 

Finally, the {\em inference strategy with global learning}
investigates the contribution of global information to
the inference model based on learning. This strategy incorporates
global information in the previous inference model in two ways. First
and most importantly,
candidate scoring is now trained online with global feedback
from the inference component. In other words, the online learning
algorithm corrects the mistakes found when comparing the correct
solution with the one generated {\em after} inference. Second, we
integrate global information in the actual inference component:
instead of performing inference for each proposition independently, we
now do it for the whole sentence at once. This allows implementation of
additional global domain constraints, e.g., arguments attached to
different predicates can not overlap. 

All the combination strategies proposed are described in detail
in Section~\ref{sec:models} and evaluated in Section~\ref{sec:results}.

\section{Individual SRL Models}
\label{sec:base}

This section introduces the three individual SRL models used by all
the combination strategies discussed in this paper. The first two
models are variations of the same algorithm: they both model the SRL
problem as a sequential tagging task, where each semantic argument is
matched to a sequence of non-embedding phrases, but Model~1 uses only
partial syntax (chunks and clause boundaries), whereas Model~2 uses
full syntax. The third model takes a more ``traditional'' approach by
assuming that there exists a one-to-one mapping between semantic
arguments and syntactic phrases. 

It is important to note that all the combination strategies introduced
later in the paper are independent of the individual SRL models
used. In fact, in Section~\ref{sec:results} we describe experiments
that use not only these individual models but also the best performing
SRL systems at the CoNLL-2005 evaluation~\cite{car05}. Nevertheless,
we choose to focus mainly on the individual SRL approaches presented
in this section for completeness and to show that state-of-the-art
performance is possible with relatively simple SRL models.  

\subsection{Models 1 and 2}

These models approach SRL as a sequential tagging task. In a
pre-processing step, the input syntactic structures are traversed in
order to select a subset of constituents organized sequentially (i.e.,
non embedding). The output of this process is a sequential
tokenization of the input sentence for each of the verb predicates.
Labeling these tokens with appropriate tags allows us to codify the
complete 
argument structure of each predicate in the sentence.

More precisely, given a verb predicate, the sequential tokens are
selected as follows: 
First, the input sentence is split into disjoint sequential {\it segments} 
using as markers for segment start/end the verb position and the 
boundaries of all the clauses that include the corresponding 
predicate constituent.
Second, for each segment,
the set of top-most non-overlapping syntactic constituents completely
falling inside the segment are selected as {\it tokens}. Finally,
these tokens are labeled with B-I-O tags, depending if they are at the
beginning, inside, or outside of a predicate argument. Note that this
strategy provides a set of sequential tokens covering the complete
sentence. Also, it is independent of the syntactic annotation
explored, assuming it provides clause boundaries.

\psfigxx{fig:seqtok1}{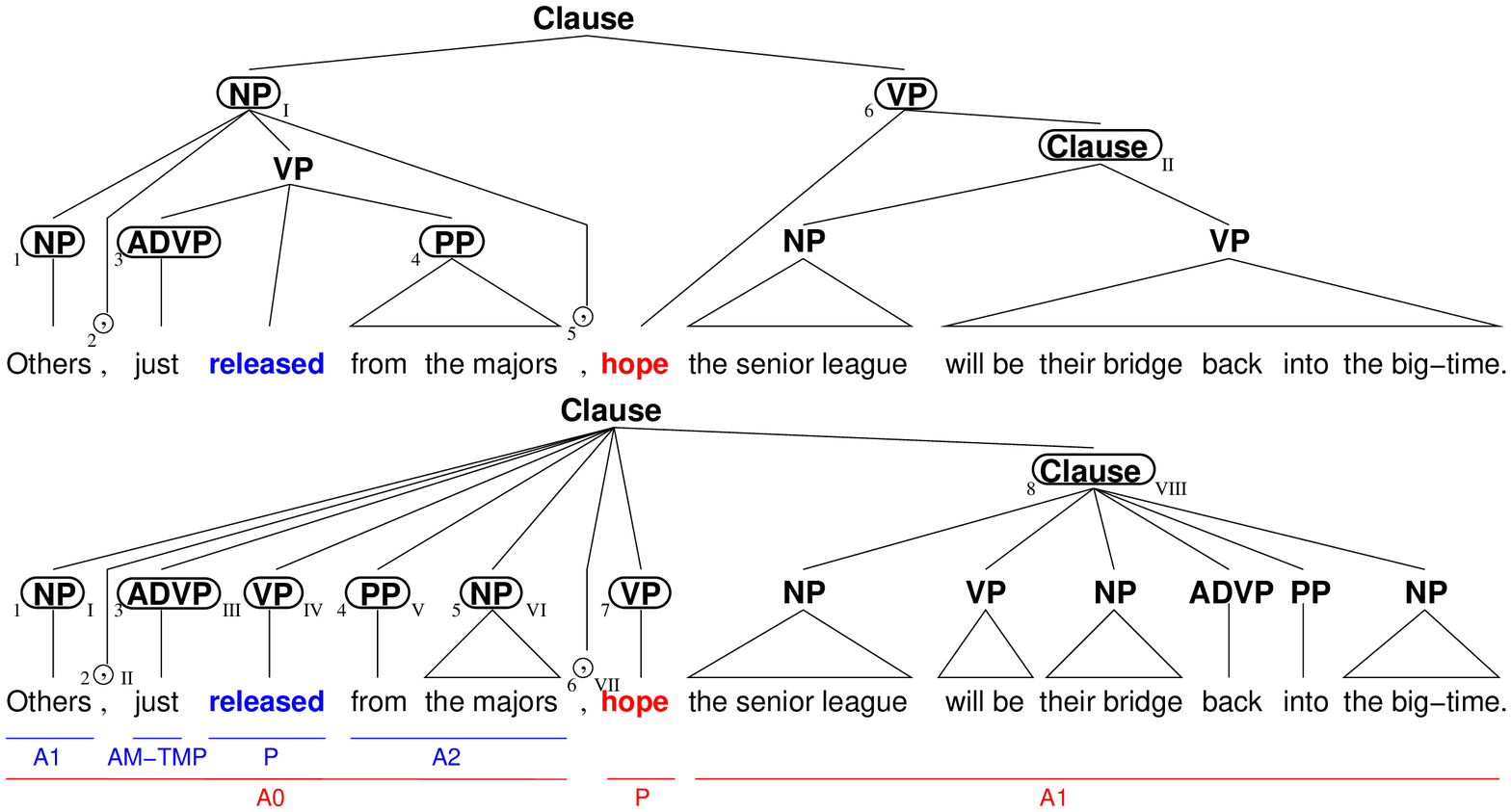}{0}
{Annotation of an example sentence with two alternative syntactic
structures. The lower tree corresponds to a partial parsing
annotation (PP) with base chunks and clause structure, while the
upper represents a full parse tree (FP). Semantic roles for two
predicates (``release'' and ``hope'') are also provided for the
sentence. The encircled nodes in both trees correspond to the selected
nodes by the process of sequential tokenization of the sentence. 
We mark the selected nodes for the predicate ``release'' with Western numerals and the nodes selected for ``hope'' with Roman numerals.
See Figure~\ref{fig:seqtok2} for more details.}{t}{5.5}

\begin{figure}[t]
\centering
{\small
\begin{tabular}{|l|l|l|l|l|}
\cline{2-5}
\multicolumn{1}{l|}{} & \multicolumn{4}{c|}{tokens} \\
\multicolumn{1}{l|}{words} & 
\multicolumn{1}{l}{{\em release}--PP} & 
\multicolumn{1}{l}{{\em release}--FP} & 
\multicolumn{1}{l}{{\em hope}--PP} & 
\multicolumn{1}{l|}{{\em hope}--FP} 
\\\hline
\phantom{1}1:\ \  Others  & 1: {\tt B\_A1} & 1: {\tt B\_A1} & \phantom{IIV}I: {\tt B\_A0} & \\
\cline{2-4}
\phantom{1}2:\ \ ,       & 2: {\tt O} & 2: {\tt O} & \phantom{IV}II: {\tt I\_A0} &\\
\cline{2-4}
\phantom{1}3:\ \ just    & 3: {\tt B\_AM-TMP} & 3: {\tt B\_AM-TMP} & \phantom{V}III: {\tt I\_A0} &\\
\cline{2-4}
\phantom{1}4:\ \ {\bf released}& \multicolumn{1}{c|}{\sf ---} & \multicolumn{1}{c|}{\sf ---} & \phantom{II}IV: {\tt I\_A0} & I: {\tt B\_A0} \\
\cline{2-4}
\phantom{1}5:\ \ from    & 4: {\tt B\_A2} & & \phantom{III}V: {\tt I\_A0} &\\
\cline{2-2}
\cline{4-4}
\phantom{1}6:\ \ the     & 5: {\tt I\_A2} & 4: {\tt B\_A2} & \phantom{II}VI: {\tt I\_A0} &\\
\phantom{1}7:\ \ majors  && &&\\
\cline{2-4}
\phantom{1}8:\ \ ,       & 6: {\tt O} & 5: {\tt O} & \phantom{I}VII: {\tt I\_A0} &\\
\cline{2-5}
\phantom{1}9:\ \ {\bf hope}    & 7: {\tt O} & & \multicolumn{1}{c|}{\sf ---} & \multicolumn{1}{c|}{\sf ---}\\
\cline{2-2}
\cline{4-5}
10:\ \ the     &&&&\\
11:\ \ senior  &&&&\\
12:\ \ league  &&&&\\
13:\ \ will    &&&&\\
14:\ \ be      && 6: {\tt O} &&\\
15:\ \ their   & 8: {\tt O} & & VIII: {\tt B\_A1} & II: {\tt B\_A1} \\
16:\ \ bridge  &&&&\\
17:\ \ back    &&&&\\
18:\ \ into    &&&&\\
19:\ \ the     &&&&\\
20:\ \ big-time&&&&\\
\hline
\end{tabular}}
\caption{Sequential tokenization of the sentence in Figure~\ref{fig:seqtok1}
according to the two syntactic views and predicates (PP stands for
partial parsing and FP for full parsing). The sentence and semantic
role annotations are vertically displayed. Each token is numbered with
the indexes that appear in the tree nodes of Figure~\ref{fig:seqtok1}
and contains the B-I-O annotation needed to codify the proper semantic
role structure.}
\label{fig:seqtok2}
\end{figure}

Consider the example in Figure~\ref{fig:seqtok1}, which depicts the
PropBank annotation of two verb predicates of a sentence (``release''
and ``hope'') and the corresponding partial and full parse trees.
Since both verbs are in the main clause of the sentence, only two
segments of the sentence are considered for both predicates, i.e.,
those defining the left and right contexts of the verbs ([w$_1$:Others,
..., w$_3$:just] and [w$_5$:from, ..., w$_{20}$:big-time] for predicate
``release'', and [w$_1$:Others, ..., w$_8$:,] and [w$_{10}$:the, ...,
w$_{20}$:big-time] for the predicate ``hope''). Figure~\ref{fig:seqtok2} shows
the resulting tokenization for both predicates and the two alternative
syntactic structures. In this case, the correct argument annotation
can be recovered in all cases, assuming perfect labeling of the tokens.

It is worth noting that the resulting number of tokens to annotate is
much lower than the number of words in all cases. Also, the
codifications coming from full parsing have substantially fewer tokens
than those coming from partial parsing. 
For example, for the predicate ``hope'', 
the difference in number of tokens 
between the two syntactic views is particularly large (8 vs. 2 tokens). 
Obviously, the coarser the token
granularity, the easier the problem of assigning correct output
labelings (i.e., there are less tokens to label and also the
long-distance relations among sentence constituents can be better
captured). On the other hand, a coarser granularity tends to introduce
more unrecoverable errors in the pre-processing stage. There is a
clear trade-off, which is difficult to solve in advance. By using the
two models in a combination scheme we can take advantage of the
diverse sentence tokenizations (see Sections~\ref{sec:models}
and~\ref{sec:results}).

Compared to the more common tree node labeling approaches (e.g., the
following Model~3), the B-I-O annotation of tokens has the advantage
of permitting to correctly annotate some arguments that do not match
a unique syntactic constituent. On the bad side, the heuristic
pre-selection of only some candidate nodes for each predicate,
i.e., the nodes that sequentially cover the sentence, 
makes the number of
unrecoverable errors higher. Another source of errors common to all
strategies are the errors introduced by real partial/full parsers. We
have calculated that due to syntactic errors introduced
in the pre-processing stage, the upper-bound
recall figures are 95.67\% for Model~1 and 90.32\% for Model~2 using
the datasets defined in Section \ref{sec:results}.

Approaching SRL as a sequential tagging task is not new. 
\citeA{hacioglu04} presented a system based on sequential tagging of
base chunks with B-I-O labels, which was the best performing SRL
system at the CoNLL-2004 shared task \cite{car04}. The novelty of our
approach resides in the fact that the sequence of syntactic tokens to
label is extracted from a hierarchical syntactic annotation (either a
partial or a full parse tree) and it is not restricted to base chunks
(i.e., a token may correspond to a complex syntactic phrase or even a
clause).

\subsubsection{Features}

Once the tokens selected are labeled with B-I-O tags, they are
  converted into training examples by considering a rich set of
  features, mainly borrowed from state-of-the-art
  systems~\cite{gil02,car04b,xue04}.  These features codify properties
  from: (a) the focus token, (b) the target predicate, 
  (c) the sentence fragment between the token and predicate, and (d)
  the dynamic context, i.e., B-I-O labels previously generated. We describe
  these four feature sets next.\footnote{Features extracted from
  partial parsing and Named Entities are common to Model~1 and 2,
  while features coming from full parse trees only apply to Model~2.}

%\newpage
{\flushleft {\bf Constituent structure features:}}
\begin{itemize}
\item Constituent {\it type} and {\it head}: extracted using the
  head-word rules of~\citeA{col99b}. If the first element is a PP
  chunk, then the head of the first NP is extracted. For example, the
  type of the constituent ``in the U.S.'' in Figure~\ref{fig:pb} is
  {\tt PP}, but its head is ``U.S.'' instead of ``in''. 
\vspace{-0.05in}
\item {\it First and last words and POS tags} of the constituent,
  e.g., ``in''/{\tt IN} and ``U.S.''/{\tt NNP} for the constituent
  ``in the U.S.'' in Figure~\ref{fig:pb}. 
\vspace{-0.05in}
\item {\it POS sequence}: if it is less than 5 tags long, e.g., {\tt
  IN$-$DT$-$NNP} for the above sample constituent. 
\vspace{-0.05in}
\item {\it 2/3/4-grams} of the POS sequence.
\vspace{-0.05in}
\item {\it Bag-of-words} of nouns, adjectives, and adverbs. For
  example, the bag-of-nouns for the constituent ``The luxury auto
  maker'' is \{``luxury'', ``auto'', ``maker''\}.  
\vspace{-0.05in}
\item {\it TOP sequence}: sequence of types of the top-most syntactic
  elements in the constituent (if it is less than 5 elements long). In
  the case of full parsing this corresponds to the right-hand side of
  the rule expanding the constituent node. For example, the TOP
  sequence for the constituent ``in the U.S.'' is {\tt IN$-$NP}.  
\vspace{-0.05in}
\item {\it 2/3/4-grams} of the TOP sequence.
\vspace{-0.05in}
\item {\it Governing category} as described by \citeA{gil02}, which
  indicates if {\tt NP} arguments are dominated by a sentence (typical
  for subjects) or a verb phrase (typical for objects). For example,
  the governing category for the constituent ``1,214 cars'' in
  Figure~\ref{fig:pb} is {\tt VP}, which hints that its corresponding
  semantic role will be object. 
\vspace{-0.05in}
\item {\it NamedEntity}, indicating if the constituent embeds or strictly
  matches a named entity along with its type. For example, the
  constituent ``in the U.S.'' embeds a locative named entity: ``U.S.''.  
\vspace{-0.05in}
\item {\it TMP}, indicating if the constituent embeds or strictly
  matches a temporal keyword (automatically extracted from {\tt
  AM-TMP} arguments of the training set). Among the most common
  temporal cue words extracted are: ``year'', ``yesterday'', ``week'',
  ``month'', etc. We used a total of 109 cue words. 
\vspace{-0.05in}
\item {\it Previous and following words and POS tag} of the
  constituent. For example, the previous word for the constituent
  ``last year'' in Figure~\ref{fig:pb} is ``maker''/{\tt NN}, and the
  next one is ``sold''/{\tt VBD}. 
\vspace{-0.05in}
\item The same features characterizing focus constituents are
  extracted for the {\it two previous and following tokens}, provided
  they are inside the boundaries of the current segment. 
\end{itemize}

{\flushleft {\bf Predicate structure features:}}
\begin{itemize}
\item Predicate {\it form}, {\it lemma}, and {\it POS tag}, e.g.,
  ``sold'', ``sell'', and {\tt VBD} for the predicate in
  Figure~\ref{fig:pb}. 
\vspace{-0.05in}
\item {\it Chunk type} and {\it cardinality of verb phrase} in which verb is
  included: single-word or multi-word. For example, the predicate in
  Figure~\ref{fig:pb} is included in a single-word {\tt VP} chunk. 
\vspace{-0.2in}
\item The predicate {\em voice}. We distinguish five voice types:
  active, passive, copulative, infinitive, and progressive. 
\vspace{-0.05in}
\item Binary flag indicating if the verb is a {\it start/end} of a clause.
\vspace{-0.05in}
\item {\em Sub-categorization rule}, i.e., the phrase structure rule
  that expands the predicate's immediate parent, e.g., {\tt S
    $\rightarrow$ NP NP VP} for the predicate in Figure~\ref{fig:pb}. 
\end{itemize}

{\flushleft {\bf Predicate-constituent features:}}
\begin{itemize}
\item {\it Relative position}, {\it distance} in words and chunks, and
  {\it level of embedding} (in number of clause-levels) with respect
  to the constituent. For example, the constituent ``in the U.S.'' in
  Figure~\ref{fig:pb} appears {\em after} the predicate, at a distance
  of 2 words or 1 chunk, and its level of embedding is 0.  
\vspace{-0.05in}
\item {\it Constituent path} as described by \citeA{gil02} and all
  {\it 3/4/5-grams} of path constituents beginning at the verb
  predicate or ending at the constituent. For example, the syntactic
  path between the constituent ``The luxury auto maker'' and the
  predicate ``sold'' in Figure~\ref{fig:pb} is {\tt NP $\uparrow$ S
    $\downarrow$ VP $\downarrow$ VBD}.  
\vspace{-0.05in}
\item {\it Partial parsing path} as described by \citeA{car04b} and all 
{\it 3/4/5-grams} of path elements beginning at the verb predicate or
ending at the constituent. For example, the path {\tt NP + PP + NP
+ S $\downarrow$ VP $\downarrow$ VBD} indicates that from
the current NP token to the predicate there are PP, NP, and S constituents
to the right (positive sign) at the same level of the token and then
the path descends
through the clause and a VP to find the predicate. The
difference from the previous constituent path is that we do not have
up arrows anymore but we introduce ``horizontal'' (left/right)
movements at the same syntactic level.
\vspace{-0.1in}
\item {\it Syntactic frame} as described by \citeA{xue04}. The
  syntactic frame captures the overall sentence structure using the
  predicate and the constituent as pivots. For example, the syntactic
  frame for the predicate ``sold'' and the constituent ``in the U.S.''
  is {\tt NP$-$NP$-${\em VP}$-$NP$-${\em PP}}, with the current
  predicate and constituent emphasized. Knowing that there are other
  noun phrases before the predicate lowers the probability that this
  constituent serves as an agent (or {\tt A0}).  
\end{itemize}

{\flushleft {\bf Dynamic features:}}
\begin{itemize}
\item {\it BIO--tag of the previous token}. When training, the correct
  labels of the left context are used. When testing, this feature is
  dynamically codified as the tag previously assigned by the SRL
  tagger.
\end{itemize}

\subsubsection{Learning Algorithm and Sequence Tagging}

We used generalized AdaBoost with real-valued weak classifiers
  \cite{sch99} as the base learning algorithm.  Our version of the
  algorithm learns fixed-depth small decision trees as weak rules,
  which are then combined in the ensemble constructed by AdaBoost.  We
  implemented a simple one-vs-all decomposition to address multi-class
  classification. In this way, a separate binary classifier has to be
  learned for each {\tt B-X} and {\tt I-X} argument label plus an extra
  classifier for the {\tt O} decision.

  AdaBoost binary classifiers are then used for {\it labeling} test
  sequences, from left to right, using a recurrent sliding window
  approach with information about the tags assigned to the preceding
  tokens. As explained in the previous list of features, left tags
  already assigned are dynamically codified as features. Empirically,
  we found that the optimal left context to be taken into account
  reduces to only the previous token.

We tested two different tagging procedures.  First, a greedy
left-to-right assignment of the best scored label for each
token. Second, a Viterbi search of the label sequence that maximizes
the probability of the complete sequence. In this case, the
classifiers' predictions were converted into probabilities using the
{\em softmax} function described in Section~\ref{sec:modelcs}. No
significant improvements were obtained from the latter. We selected
the former, which is faster, as our basic tagging algorithm for the
experiments.

Finally, this tagging model enforces three basic constraints: (a) the
B-I-O output labeling must codify a correct structure; (b) arguments
cannot overlap with clause nor chunk boundaries; and (c) for each
verb, {\tt A0-5} arguments not present in PropBank frames (taking the
union of all rolesets for the different verb senses) are not
considered.

%%%%%%%%%%%%%%%%%%%%%%%%%%%%%%%%%%%%%%%%%%%

\subsection{Model~3}
\label{sec:model3}
%\vspace*{3mm}

The third individual SRL model makes the
strong assumption that each predicate argument maps to one syntactic
constituent. For example, in Figure~\ref{fig:pb} {\tt A0} maps to a
noun phrase, {\tt AM-LOC} maps to a prepositional phrase, etc.
This assumption holds well on hand-corrected parse trees and
simplifies significantly the SRL process because only one syntactic
constituent has to be correctly classified in order to recognize one
semantic argument. On the other hand, this approach is limited when
using automatically-generated syntactic trees. For example, only
91.36\% of the arguments can be mapped to one of the
syntactic constituents produced by the Charniak parser.

Using a bottom-up approach, Model~3 maps each argument to the first
syntactic constituent that has the exact same boundaries and then
climbs as high as possible in the tree across unary production
chains. We currently ignore all arguments that do not map to a single
syntactic constituent. 
The argument-constituent mapping is performed on the training set as preprocessing step.
Figure~\ref{fig:pb} shows a mapping example
between the semantic arguments of one verb and the corresponding
sentence syntactic structure. 

Once the mapping process completes, Model~3 extracts a rich set of
lexical, syntactic, and semantic features. Most of these features are
inspired from previous work in parsing and
SRL~\cite{col99b,gil02,sur03,pra05}. We describe the complete feature
set implemented in Model~3 next. 

\subsubsection{Features}

Similarly to Models 1 and 2 we group the features in three
categories, based on the properties they codify: (a) the argument
constituent, (b) the target predicate, and (c) the relation between
the constituent and predicate syntactic constituents. 

{\flushleft {\bf Constituent structure features:}}
\begin{itemize}
\item The {\em syntactic label} of the candidate constituent.
\vspace{-0.05in}
\item The constituent {\em head word}, {\em suffixes} of length 2, 3,
  and 4, {\em lemma}, and {\em POS tag}. 
\vspace{-0.05in}
\item The constituent {\em content word}, {\em suffixes} of length 2,
  3, and 4, {\em lemma}, {\em POS tag}, and {\em NE label}. Content
  words, which add informative lexicalized information different from
  the head word, were detected using the heuristics
  of~\citeA{sur03}. For example, the head word of the verb phrase
  ``had placed'' is the auxiliary verb ``had'', whereas the content
  word is ``placed''. Similarly, the content word of prepositional
  phrases is not the preposition itself (which is selected as the head
  word), but rather the head word of the attached phrase, e.g.,
  ``U.S.'' for the prepositional phrase ``in the U.S.''.  
\vspace{-0.05in}
\item The {\em first and last constituent words} and their {\em POS tags}.
\vspace{-0.05in}
\item {\em NE labels} included in the candidate phrase. 
\vspace{-0.05in}
\item Binary features to indicate the presence of {\em temporal cue
  words}, i.e., words that appear often in {\tt AM-TMP} phrases in
  training. We used the same list of temporal cue words as Models 1
  and 2. 
\vspace{-0.05in}
\item For each Treebank syntactic label we added a feature to indicate
  the {\em number of such labels} included in the candidate phrase. 
\vspace{-0.05in}
\item The {\em TOP sequence} of the constituent (constructed similarly
  to Model~2). 
\vspace{-0.05in}
\item The phrase {\em label}, {\em head word and POS tag} of the
  constituent parent, left sibling, and right sibling. 
\end{itemize}

{\flushleft {\bf Predicate structure features:}}
\begin{itemize} 
\item The predicate {\em word} and {\em lemma}.
\vspace{-0.05in}
\item The predicate {\em voice}. Same definition as Models 1 and 2.
\vspace{-0.05in}
\item A binary feature to indicate if the predicate is {\em frequent}
  (i.e., it appears more than twice in the training data) or not. 
\vspace{-0.05in}
\item {\em Sub-categorization rule}. Same definition as Models 1 and 2.
\end{itemize}

{\flushleft {\bf Predicate-constituent features:}}
\begin{itemize}
\item The {\em path} in the syntactic tree between the argument phrase
  and the predicate as a chain of syntactic labels along with the
  traversal direction (up or down). It is computed similarly to Model
  2. 
\vspace{-0.05in}
\item The {\em length} of the above syntactic path.
\vspace{-0.05in}
\item The {\em number of clauses} ({\tt S*} phrases) in the path. We
  store the overall clause count and also the number of clauses in the
  ascending and descending part of the path.  
\vspace{-0.05in}
\item The {\em number of verb phrases} ({\tt VP}) in the
  path. Similarly to the above feature, we store three numbers:
  overall verb count, and the verb count in the ascending/descending
  part of the path. 
\vspace{-0.10in}
\item {\em Generalized syntactic paths}. We generalize the 
  path in the syntactic tree, when it appears with more than 3 
  elements, using two templates:  
(a) {\tt Arg} $\uparrow$ {\tt Ancestor} $\downarrow$ {\tt N}$_i$
  $\downarrow$   {\tt Pred}, where {\tt Arg} is the argument label,
  {\tt Pred} is the predicate label, {\tt Ancestor} is the label of
  the common ancestor, and {\tt N}$_i$ is instantiated with each of
  the labels between {\tt Pred} and {\tt Ancestor} in the full path;
  and 
(b) {\tt Arg} $\uparrow$ {\tt N}$_i$ $\uparrow$ {\tt Ancestor}
  $\downarrow$ {\tt Pred}, where {\tt N}$_i$ is instantiated with each
  of the labels between {\tt Arg} and {\tt Ancestor} in the full path.
For example, in the path {\tt NP} $\uparrow$ {\tt S} $\downarrow$ {\tt
  VP} $\downarrow$ {\tt SBAR} $\downarrow$ {\tt S} $\downarrow$ {\tt
  VP} the argument label is the first {\tt NP}, the predicate label is
  the last {\tt VP}, and the common ancestor's label is the first {\tt
  S}. Hence, using the last template, this path is generalized to the
  following three features: {\tt NP} $\uparrow$ {\tt S} $\downarrow$
  {\tt VP} $\downarrow$ {\tt VP}, {\tt NP} $\uparrow$ {\tt S}
  $\downarrow$ {\tt SBAR} $\downarrow$ {\tt VP}, and {\tt NP}
  $\uparrow$ {\tt S} $\downarrow$ {\tt S} $\downarrow$ {\tt VP}. This
  generalization reduces the sparsity of the complete
  constituent-predicate path feature using a different strategy than
  Models 1 and 2, which implement a $n$-gram based approach. 
\vspace{-0.05in}
\item The {\em subsumption count}, i.e., the difference between the
  depths in the syntactic tree of the argument and predicate
  constituents. This value is 0 if the two phrases share the same
  parent. 
\vspace{-0.05in}
\item The {\em governing category}, similar to Models~1 and 2.
%  which indicates if {\tt NP}
%  arguments are dominated by a sentence (typical for subjects) or a
%  verb phrase (typical for objects)~\cite{gil02}.  
\vspace{-0.10in}
\item The {\em surface distance} between the predicate and the
  argument phrases encoded as: the number of tokens, verb terminals
  ({\tt VB*}), commas, and coordinations ({\tt CC}) between the
  argument and predicate phrases, and a binary feature to indicate if
  the two constituents are adjacent.  
For example, the surface distance between the argument candidate
  ``Others'' and the predicate ``hope'' in the Figure~\ref{fig:seqtok1}
  example: ``Others, just released from the majors, hope the senior
  league...'' is 7 tokens, 1 verb, 2 commas, and 0 coordinations.  
These features, originally proposed by \citeA{col99b} for his
  dependency parsing model, capture robust, syntax-independent
  information about the sentence structure. For example, a constituent
  is unlikely to be the argument of a verb if another verb appears
  between the two phrases.  
\vspace{-0.05in}
\item A binary feature to indicate if the argument {\em starts with a}
  {\em predicate particle}, i.e., a token seen with the {\tt RP*} POS
  tag and directly attached to the predicate in training. The
  motivation for this feature is to avoid the inclusion of predicate
  particles in the argument constituent. For example, without this
  feature, a SRL system will tend to incorrectly include the predicate
  particle in the argument for the text: ``take [$_{\tt A1}$over the
    organization]'', because the marked text is commonly incorrectly
  parsed as a prepositional phrase and a large number of prepositional
  phrases directly attached to a verb are arguments for the
  corresponding predicate. 
\end{itemize}

\subsubsection{Classifier}

Similarly to Models 1 and 2, Model~3 trains one-vs-all classifiers
using AdaBoost for the most common argument labels. To reduce the
sample space, Model~3 selects training examples (both positive and
negative) only from: (a) the first clause that includes the predicate,
or (b) from phrases that appear to the left of the predicate in the
sentence. More than 98\% of the argument constituents fall into one of
these classes.

At prediction time the classifiers are combined using a simple greedy
technique that iteratively assigns to each predicate the argument
classified with the highest confidence. For each predicate we consider
as candidates all {\tt AM} attributes, but only numbered attributes
indicated in the corresponding PropBank frame. Additionally, this
greedy strategy enforces a limited number of domain knowledge
constraints in the generated solution: (a) arguments can not overlap
in any form, (b) no duplicate arguments are allowed for {\tt A0-5},
and (c) each predicate can have numbered arguments, i.e., {\tt A0-5},
only from the subset present in its PropBank frame. 
These constraints are somewhat different from the constraints used by
Models 1 and 2: (i) Model~3 does not use the B-I-O representation
hence the constraint that the B-I-O labeling be correct does not
apply; and (ii) Models 1 and 2 do not enforce the constraint that
numbered arguments can not be duplicated because its implementation is
not straightforward in this architecture.

\section{Performance of the Individual Models}
\label{sec:indperf}

In this section we analyze the performance of the three individual SRL
models proposed.
Our three SRL systems were trained  using the complete
CoNLL-2005 training set (PropBank/Treebank sections 2 to 21). To avoid
the overfitting of the syntactic processors $-$i.e., part-of-speech
tagger, chunker, and Charniak's full parser$-$ we partitioned the
PropBank training set into five folds and for each fold we used the
output of the syntactic processors that were trained on the other four
folds. The models were tuned on a separate development partition
(Treebank section 24) and evaluated on two corpora: (a) Treebank
section 23, which consists of Wall Street Journal (WSJ) documents, and
(b) on three sections of the Brown corpus, semantically annotated by
the PropBank team for the CoNLL-2005 shared task evaluation. 

%\subsection{Experimental Results}

All the classifiers for our individual models were
developed using AdaBoost with decision trees of depth 4 (i.e., each
branch may represent a conjunction of at most 4 basic features). Each
classification model was trained for up to 2,000 rounds.
We applied some simplifications to keep training times and memory
requirements inside admissible bounds: (a) we have
trained only the most frequent argument labels: top 41 for Model~1,
top 35 for Model~2, and top 24 for Model~3; (b) we discarded all
features occurring less than 15 times in the training set, and
(c) for each Model~3 classifier, we have limited the number of
negative training samples to the first 500,000 negative samples
extracted in the PropBank traversal\footnote{The distribution of
  samples for the Model~3 classifiers is very biased towards negative
  samples because, in the worst case, any syntactic constituent in the
  same sentence with the predicate is a potential argument.}. 

Table~\ref{tab:base} summarizes the results of the three models on the
WSJ and Brown corpora. We include the percentage of
perfect propositions detected by each model (``PProps''), i.e.,
predicates recognized with all their arguments, the overall precision,
recall, and F$_1$ measure\footnote{The significance intervals
for the F$_1$ measure have been obtained using bootstrap resampling
\cite{nor89}. F$_1$ rates outside of these intervals are assumed to be
significantly different from the related F$_1$ rate ($p<0.05$).}.
The results summarized in Table~\ref{tab:base} indicate that all
individual systems have a solid performance. Although none of them
would rank in the top 3 in the CoNLL-2005 evaluation~\cite{car05},
their performance is comparable to the best individual systems
presented at that evaluation exercise\footnote{The best performing
SRL systems at CoNLL were a combination of several subsystems. See
section \ref{sec:related} for details.}.
Consistently with other systems evaluated on the Brown
corpus, all our models experience a severe performance drop in this
corpus, due to the lower performance of the linguistic processors.

\begin{table}[t]
%\begin{small}
\begin{center}
\begin{tabular}{|l|c|c|c|c|}\cline{2-5}
\multicolumn{1}{l|}{WSJ} & PProps & Precision & Recall &
F$_1$\\
\hline
Model 1  & 48.45\% & 78.76\% & 72.44\% & 75.47 {\scriptsize$\pm 0.8$}\\
Model 2  & {\bf 52.04}\% & 79.65\% & {\bf 74.92}\% & {\bf 77.21} {\scriptsize$\pm 0.8$}\\
Model 3  & 45.28\% & {\bf 80.32}\% & 72.95\% & 76.46 {\scriptsize$\pm 0.6$}\\
\hline
%\end{tabular}
%\begin{tabular}{|l|c|c|c|c|}%\cline{2-5}
\multicolumn{5}{l}{Brown}\\
\hline
Model 1  & 30.85\% & 67.72\% & 58.29\% & 62.65 {\scriptsize$\pm 2.1$}\\
Model 2  & {\bf 36.44}\% & 71.82\% & {\bf 64.03}\% & {\bf 67.70} {\scriptsize$\pm 1.9$}\\
Model 3  & 29.48\% & {\bf 72.41}\% & 59.67\% & 65.42 {\scriptsize$\pm 2.1$}\\
\hline
\end{tabular}
%\vspace{-0.1in}
%\end{small}
\end{center}
\caption{Overall results of the individual models in the WSJ and Brown
  test sets.}
\label{tab:base}
\end{table}

As expected, the models based on full parsing (2 and 3) perform better
than the model based on partial syntax. But, interestingly, the
difference is not large (e.g., less than 2 points in
F$_1$ in the WSJ corpus), evincing that having base
syntactic chunks and clause boundaries is enough to obtain
competitive performance. More importantly, the full-parsing models are
{\em not always} better than the partial-syntax
model. Table~\ref{tab:baseargs} lists the F$_1$ measure for
the three models for the first five numbered
arguments. Table~\ref{tab:baseargs} shows that Model~2, our overall
best performing individual system, achieves the best F-measure for
{\tt A0} and {\tt A1} (typically subjects and direct objects), but
Model~1, the partial-syntax model, performs best for the {\tt A2}
(typically indirect objects, instruments, or benefactives). The
explanation for this behavior is that indirect objects tend to be
farther from their predicates and accumulate more parsing errors. 
From the models based on full syntax, Model~2 has better recall whereas
Model~3 has better precision, because Model~3 filters out {\em all} candidate
arguments that do not match a single syntactic constituent.
Generally, Table~\ref{tab:baseargs} shows that all models have strong
and weak points. This is further justification for our focus on
combination strategies that combine several independent models. 

\begin{table}[t]
\centering
%\begin{small}
%\begin{tabular}{|l|c|c|c|}\cline{2-4}
%\multicolumn{1}{l|}{} & M1 F$_1$ & M2 F$_1$ & M3 F$_1$ \\
%\hline
%{\tt A0}& 83.37 & {\bf 86.65} & 86.14 \\
%{\tt A1}& 75.13 & {\bf 77.06} & 75.83 \\
%{\tt A2}& {\bf 67.33} & 65.04 & 65.55 \\
%{\tt A3}& 61.92 & 62.72 & {\bf 65.26} \\
%{\tt A4}& 72.73 & 72.43 & {\bf 73.85} \\
%\hline
%\begin{small}
\begin{tabular}{|l|c|c|c|c|c|}\cline{2-6}
\multicolumn{1}{l|}{} & {\tt A0} &  {\tt A1} & {\tt A2} & {\tt A3} & {\tt A4}\\
\hline
Model 1 F$_1$ & 83.37 & 75.13 & {\bf 67.33} & 61.92 & 72.73 \\
Model 2 F$_1$ & {\bf 86.65} & {\bf 77.06} & 65.04 & 62.72 & 72.43 \\
Model 3 F$_1$ & 86.14 & 75.83 & 65.55 & {\bf 65.26} & {\bf 73.85} \\
\hline
\end{tabular}
%\end{small}
\caption{F$_{1}$ scores of the individual systems for the {\tt A0}$-${\tt 4} arguments in the WSJ test.}
\label{tab:baseargs}
\end{table}

\section{Features of the Combination Models}
\label{sec:features}

As detailed in Section~\ref{sec:approach}, in this paper we analyze
two classes of combination strategies for the problem of semantic role
labeling: (a) an inference model with constraint satisfaction, which
finds the set of candidate arguments that maximizes a global cost
function, and (b) two inference strategies based on learning, where
candidates are scored and ranked using discriminative
classifiers. From the perspective of the feature space, the main
difference between these two types of combination models is that the
input of the first combination strategy is limited to the argument
probabilities produced by the individual systems, whereas the last
class of combination approaches incorporates a much larger feature set
in their ranking classifiers. For robustness, in this paper we use
only features that are extracted from the solutions provided by the
individual systems, hence are independent of the individual
models\footnote{With the exception of the argument probabilities,
  which are required by the constraint satisfaction model.}. 
We describe all these features next. All examples given in this section are based on Figures~\ref{fig:combo} and~\ref{fig:chunks}.

{\flushleft {\bf Voting features}} $-$ these features quantify the
votes received by each argument from the individual systems. This set
includes the following features: 
\begin{itemize}
\item The {\em label} of the candidate argument, e.g., {\tt A0} for
  the first argument proposed by system M1 in Figure~\ref{fig:combo}. 
\vspace{-0.05in}
\item The {\em number of systems} that generated an argument with this
  label and span. For the example shown in Figure~\ref{fig:combo},
  this feature has value 1 for the argument {\tt A0} proposed by M1
  and 2 for M1's {\tt A1}, because system M2 proposed the same argument.
\vspace{-0.05in}
\item The {\em unique ids} of all the systems that generated an
  argument with this label and span, e.g., M1 and M2 for the argument
  {\tt A1} proposed by M1 or M2 in Figure~\ref{fig:combo}. 
\vspace{-0.05in}
\item The {\em argument sequence} for this predicate for all the
  systems that generated an argument with this label and span. For
  example, the argument sequence generated by system M1 for the
  proposition illustrated in Figure~\ref{fig:combo} is: {\tt A0 - V -
    A1 - A2}. This feature attempts to capture information at
  proposition level, e.g., a combination model might learn to trust
  model M1 more for the argument sequence {\tt A0 - V - A1 - A2}, M2
  for another sequence, etc. 
\end{itemize}

\psfigxx{fig:combo}{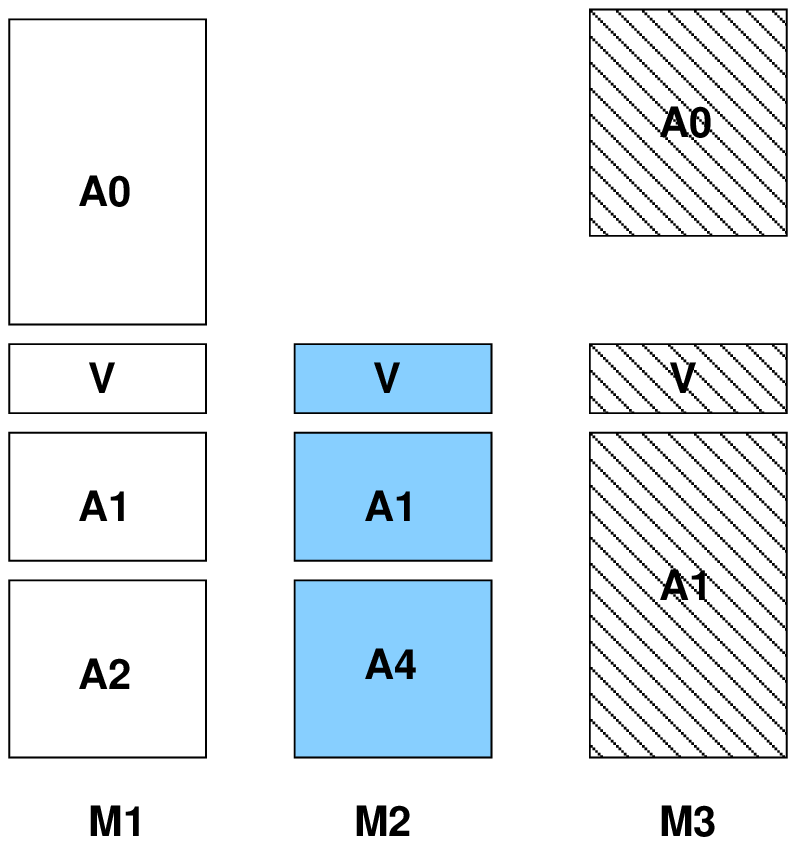}{0.15}{Sample solutions proposed for the
  same predicate by three individual SRL models: M1, M2 and
  M3. Argument candidates are displayed vertically for each system.}{tp}{2.5}

{\flushleft {\bf Same-predicate overlap features}} $-$ these features
measure the overlap between different arguments produced by the
individual SRL models for the {\em same} predicate: 
\begin{itemize}
\item The {\em number} and {\em unique ids} of all the systems that
  generated an argument with the {\em same span but different
    label}. For the example shown in Figure~\ref{fig:combo}, these
  features have values 1 and M2 for the argument {\tt A2} proposed by
  M1, because model M2 proposed argument {\tt A4} with the same span. 
\vspace{-0.05in}
\item The {\em number} and {\em unique ids} of all the systems that
  generated an argument {\em included} in the current argument. For
  the candidate argument {\tt A0} proposed by model M1 in
  Figure~\ref{fig:combo}, these features have values 1 and M3, because
  M3 generated argument {\tt A0}, which is included in M1's {\tt A0}. 
\vspace{-0.05in}
\item In the same spirit, we generate the {\em number} and {\em unique
  ids} of all the systems that generated an argument that {\em
  contains} the current argument, and the {\em number} and {\em unique
  ids} of all the systems that generated an argument that {\em
  overlaps} $-$ but does not include nor contain $-$ the current
  argument. 
\end{itemize}

{\flushleft {\bf Other-predicate overlap features}} $-$  these
features quantify the overlap between different arguments produced by
the individual SRL models for {\em other} predicates. We generate the
same features as the previous feature group, with the difference that
we now compare arguments generated for different predicates. The
motivation for these overlap features is that, according to the
PropBank annotations, no form of overlap is allowed among arguments
attached to the same predicate, and only inclusion or containment is
permitted between arguments assigned to different predicates. The
overlap features are meant to detect when these domain constraints are
not satisfied by a candidate argument, which is an indication, if the
evidence is strong, that the candidate is incorrect. 

{\flushleft {\bf Partial-syntax features}} $-$ these features codify
the structure of the argument and the distance between the argument
and the predicate using only partial syntactic information, i.e.,
chunks and clause boundaries (see Figure~\ref{fig:chunks} for an
example). Note that these features are inherently different from the
features used by Model~1, because Model~1 evaluates each individual
chunk part of a candidate argument, whereas here we codify properties
of the complete argument constituent. We describe the partial-syntax
features below. 
\begin{itemize}
\item {\em Length in tokens and chunks} of the argument constituent,
  e.g., 4 and 1 for argument {\tt A0} in Figure~\ref{fig:chunks}. 
\vspace{-0.05in}
\item The {\em sequence of chunks included in the argument
  constituent}, e.g., {\tt PP NP} for the argument {\tt AM-LOC} in
  Figure~\ref{fig:chunks}. If the chunk sequence is too large, we
  store $n$-grams of length 10 for the start and end of the sequence.  
\vspace{-0.05in}
\item The {\em sequence of clause boundaries}, i.e., clause beginning
  or ending, {\em included in the argument constituent}. 
\vspace{-0.05in}
\item The {\em named entity types} included in the argument
  constituent, e.g., {\tt LOCATION} for the {\tt AM-LOC} argument in
  Figure~\ref{fig:chunks}. 
\vspace{-0.05in}
\item {\em Position of the argument}: before/after the predicate in
  the sentence, e.g., {\em after} for {\tt A1} in
  Figure~\ref{fig:chunks}. 
\vspace{-0.05in}
\item A Boolean flag to indicate if the argument constituent is {\em
  adjacent} to the predicate, e.g., {\em false} for {\tt A0} and {\em
  true} for {\tt A1} in Figure~\ref{fig:chunks}. 
\vspace{-0.05in}
\item The {\em sequence of chunks between the argument constituent and
  the predicate}, e.g., the chunk sequence between the predicate and
  the argument {\tt AM-LOC} in Figure~\ref{fig:chunks} is: {\tt
    NP}. Similarly to the above chunk sequence feature, if the
  sequence is too large, we store starting and ending $n$-grams. 
\vspace{-0.05in}
\item The {\em number of chunks between the predicate and the
  argument}, e.g., 1 for {\tt AM-LOC} in Figure~\ref{fig:chunks}. 
\vspace{-0.05in}
\item The {\em sequence of clause boundaries between the argument
  constituent and the predicate}.  
\vspace{-0.2in}
\item The {\em clause subsumption count}, i.e., the difference between
  the depths in the clause tree of the argument and predicate
  constituents. This value is 0 if the two phrases are included in the
  same clause. 
\end{itemize}

\psfigxx{fig:chunks}{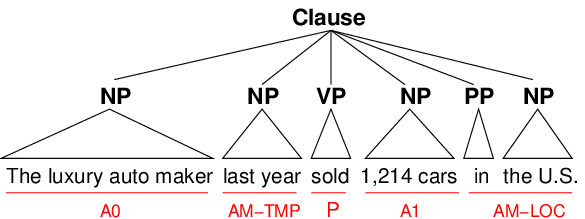}{0.15}{Sample proposition with partial
  syntactic information.}{tp}{3.5} 

{\flushleft {\bf Full-syntax features}} $-$ these features codify the
structure of the argument constituent, the predicate, and the distance
between the two using full syntactic information. The full-syntax
features are replicated from Model~3 (see Section~\ref{sec:model3}),
which assumes that a one-to-one mapping from semantic constituents to
syntactic phrases exists. Unlike Model~3 which ignores arguments that
can not be matched against a syntactic constituent, if such an exact
mapping does not exist due to the inclusion of candidates from Models 1 and 2,
we generate an approximate mapping from the
unmapped semantic constituent to the largest phrase that is included
in the given span and has the same left boundary as the semantic
constituent. This heuristic guarantees that we capture at least some
of the semantic constituents' syntactic structure. 

The motivation for the partial and full-syntax features is to learn
the ``preferences'' of the individual SRL models. For example, with
these features a combination classifier might learn to trust model M1
for arguments that are closer than 3 chunks to the predicate, model M2
when the predicate-argument syntactic path is {\tt NP} $\uparrow$ {\tt
  S} $\downarrow$ {\tt VP} $\downarrow$ {\tt SBAR} $\downarrow$ {\tt
  S} $\downarrow$ {\tt VP}, etc. 

{\flushleft {\bf Individual systems' argument probabilities}} $-$ each
individual model outputs a confidence score for each of their proposed
arguments. These scores are converted into probabilities
using the \emph{softmax} function as described in detail in Section
\ref{sec:modelcs}. The combination strategy based on constraint
satisfaction (Section \ref{sec:modelcs}) uses these probabilities as
they are, while the other two strategies based on meta-learning
(Section \ref{sec:modelwofeed}) have to
discretize the probabilities to include them as features. To do so,
each probability value is matched to one of five probability intervals
and the corresponding interval is used as the feature. The probability
intervals are dynamically constructed for each argument label and each
individual system such that the corresponding system predictions for
this argument label are uniformly distributed across the intervals.  

In Section~\ref{sec:perfwofeed} we empirically analyze the
contribution of each of these proposed feature sets to the performance
of our best combination model. 

\section{Combination Strategies}
\label{sec:models}

In this section we detail the combination strategies proposed in this
paper: (a) a combination model with constraint satisfaction, which
aims at finding the set of candidate arguments that maximizes a global
cost function, and (b) two combination models with inference based on
learning, where candidates are scored and ranked using discriminative
classifiers.  
In the previous section we described the complete feature set made
available to all approaches. Here we focus on the machine learning
paradigm deployed by each of the combination models. 

\subsection{Inference with Constraint Satisfaction}
\label{sec:modelcs}

The Constraint Satisfaction model selects a subset of candidate
arguments that maximizes a compatibility function subject to
the fulfillment of a set of structural constraints that ensure
consistency of the solution. The compatibility function is based on
the probabilities given by individual SRL models to the candidate
arguments. In this work we use Integer Linear
Programming to solve the constraint satisfaction problem. This
approach was first proposed by \citeA{roth04a} and applied to semantic
role labeling by \citeA{roth04b,koo05}, among others. We follow the
setting of~Komen et al., which is taken as a reference.

As a first step, the scores from each model are normalized into
probabilities. The scores yielded by the classifiers are signed and
unbounded real numbers, but experimental evidence shows that the
confidence in the predictions (taken as the absolute value of the raw
scores) correlates well with the classification accuracy.  Thus, the
\emph{softmax} function \cite{bish95} is used to convert the set of
unbounded scores into probabilities. If there are $k$ possible output
labels for a given argument and ${\tt sco}(l_i)$ denotes the score of
label $l_i$ output by a fixed SRL model, then the estimated probability
for this label is:

\begin{displaymath}
\hat p(l_i) = \frac{e^{\gamma {\tt sco}(l_i)}}{\sum^{k}_{j=1}{e^{\gamma {\tt sco}(l_j)}}}
\end{displaymath}

The $\gamma$ parameter of the above formula can be empirically
adjusted to avoid overly skewed probability distributions and to
normalize the scores of the three individual models to a similar range
of values. See more details about our experimental setting in
Section~\ref{sec:expsettings}.

Candidate selection is performed via Integer Linear Programming
(ILP). The program goal is to maximize a compatibility function
modeling the global confidence of the selected set of candidates,
subject to a set of linear constraints. All the variables
involved in the task take integer values and may appear in first
degree polynomials only.

An abstract ILP process can be described in a simple fashion as: given
a set of variables $V=\{v_1,\dots, v_n\}$, it aims to maximize the
global compatibility of a label assignment $\{l_1,\ldots,l_n\}$ to
these variables.  A local compatibility function $c_v(l)$ defines the
compatibility of assigning label $l$ to variable $v$. The global
compatibility function $C(l_1,\ldots,l_n)$ is taken as the sum of each
local assignment compatibility, so the goal of the ILP process can be
written as:

\begin{displaymath}
 \begin{array}{c}
   \phantom{1} \\
   \mathrm{argmax} \\
   ^{l_1,\ldots,l_n}
 \end{array}
  C(l_1,\ldots,l_n) = 
 \begin{array}{c}
   \phantom{1} \\
   \mathrm{argmax} \\
   ^{l_1,\ldots,l_n}
 \end{array}
  \sum_{i = 1}^{n}c_{v_i}(l_i)
\end{displaymath}

\noindent where the constraints are described in a set of accompanying 
integer linear equations involving the variables of the problem.

If one wants to codify soft constraints instead of hard, there is the
possibility of considering them as a penalty component in the
compatibility function.  In this case, each constraint $r\in R$ can be
seen as a function which takes the current label assignment and
outputs a real number, which is 0 when the constraint is satisfied and
a positive number when not, indicating the penalty imposed to the
compatibility function. The new expression of the compatibility
function to maximize is:

\begin{displaymath}
  C(l_1,\ldots,l_n) = \sum_{i=1}^{n}c_{v_i}(l_i) - \sum_{r\in R} r(l_1,\ldots,l_n)
\end{displaymath}

\noindent Note that the hard constraints can also be simulated in this setting
by making them output a very large positive number when they are
violated.

In our particular problem, we have a binary-valued variable $v_i$ for
each of the $N$ argument candidates generated by the SRL models, i.e.,
$l_i$ labels are in $\{0,1\}$. Given a label assignment, the arguments
with $l_i=1$ are selected to form the solution, while the others
(those where $l_i=0$) are filtered out. For each variable
$v_i$, we also have the probability values, $p_{ij}$, calculated from
the score of model $j$ on argument {i}, according to the {\em softmax}
formula described above\footnote{If model $j$ does not propose
argument $i$ then we consider $p_{ij}=0$.}. In a first approach, the
compatibility function $c_v(l_i)$ equals to
$(\sum_{j=1}^{M}p_{ij})l_i$, where the number of models, $M$, is 3 in
our case\footnote{Instead of accumulating the probabilities of all
models for a given candidate argument, one could consider a different
variable for each model prediction and introduce a constraint forcing
all these variables to take the same value at the end of the
optimization problem. The two alternatives are equivalent.}.

Under this definition, maximizing the compatibility function is
equivalent to maximizing the sum of the probabilities given by the
models to the argument candidates considered in the solution.  Since
this function is always positive, the global score increases directly
with the number of selected candidates. As a consequence, the model is
biased towards the maximization of the number of candidates included
in the solution (e.g., tending to select a lot of small
non-overlapping arguments). Following \citeA{koo05}, this bias can be
corrected by adding a new score $o_i$, which sums to the compatibility
function when the $i$-th candidate is not selected in the
solution. The global compatibility function needs to be rewritten to
encompass this new information. Formalized as an ILP equation, it looks
like:

\begin{displaymath}
 \begin{array}{c}
   \phantom{1} \\
   \mathrm{argmax} \\
   ^{L \in \{0,1\}^{N}}
 \end{array}
  C(l_1,\ldots,l_N) = 
 \begin{array}{c}
   \phantom{1} \\
   \mathrm{argmax} \\
   ^{L \in \{0,1\}^{N}}
 \end{array}
 \sum_{i=1}^{N} (\sum_{j=1}^{M}p_{ij})l_{i}+o_{i}(1-l_{i})
\end{displaymath}

\noindent where the constraints are expressed in separated integer linear
equations. It is not possible to define \emph{a priori} the value of
$o_i$. Komen et al. used a validation corpus to empirically estimate
a constant value for all $o_i$ (i.e., independent from the argument
candidate)\footnote{Instead of working with a constant, one could try
to set the $o_i$ value for each candidate, taking into account some
contextual features of the candidate. We plan to explore this option
in the near future.}. We will use exactly the same solution
of working with a single constant value, to which we will refer as
$O$.

Regarding the consistency constraints, we have considered the following six:
%\vspace{5mm}
\begin{enumerate}
\item Two candidate arguments for the same verb can not overlap nor
  embed.
  \vspace{-8pt}
\item A verb may not have two core arguments with the same type label
  {\tt A0}-{\tt A5}.
  \vspace{-8pt}
\item If there is an argument {\tt R-X} for a verb, there has to be
  also an {\tt X} argument for the same verb.
  \vspace{-8pt}
\item If there is an argument {\tt C-X} for a verb, there has to be
  also an {\tt X} argument before the {\tt C-X} for the same verb.
  \vspace{-8pt}
\item Arguments from two different verbs can not overlap, but they
  can embed.
  \vspace{-8pt}
\item Two different verbs can not share the same {\tt AM-X}, {\tt
  R-AM-X} or {\tt C-X} arguments.
\end{enumerate}

Constraints 1--4 are also included in our reference work
\cite{roth04b}. No other constraints from that paper need to be
checked here since each individual model outputs only consistent
solutions.  Constraints 5 and 6, which 
restrict the set of compatible arguments among different
predicates in the sentence, are original to this work.
In the Integer Linear Programming setting
the constraints are written as inequalities. For example, if $A_i$ is
the argument label of the $i$-th candidate and $V_i$ its verb
predicate, constraint number $2$ is written as: $\sum_{(A_{i} = a
\,\land\, V_{i} = v)}\, l_{i} \le 1$, for a given verb $v$ and argument 
label $a$. The other constraints have similar translations into
inequalities.

Constraint satisfaction optimization will be applied in two different
ways to obtain the complete output annotation of a sentence. In the
first one, we proceed verb by verb independently to find their best
selection of candidate arguments using only constraints 1 through
4. We call this approach {\em local} optimization. In the second
scenario all the candidate arguments in the sentence are considered at
once and constraints 1 through 6 are enforced. We will refer to this
second strategy as {\em global} optimization. In both scenarios the
compatibility function will be the same, but constraints need some
rewriting in the global scenario because they have to include
information about the concrete predicate.

In Section \ref{sec:perfcs} we will extensively evaluate the presented
inference model based on Constraint Satisfaction, and we will describe
some experiments covering the following topics: (a) the
contribution of each of the proposed constraints; (b) the
performance of {\em local} vs. {\em global} optimization; and (c) 
the precision--recall tradeoff by varying the value of the
bias-correction parameter.

%%%%%%%%%%%%%%%%%%%%%%%%%%%%%%%%
% DESCRIPTION OF COMBINATION MODELS BASED ON LEARNING

\subsection{Inference Based On Learning}
\label{sec:modelwofeed}

This combination model consists of two stages: a {\em candidate
  scoring} phase, which scores candidate arguments in the pool using a
series of discriminative classifiers, and an {\em inference} stage,
which selects the best overall solution that is consistent with the
domain constraints.

The first and most important component of this combination strategy is
the candidate scoring module, which assigns to each candidate argument 
a score equal to the confidence that this argument is part of the global
solution. 
It is formed by discriminative functions, one for each role label.
Below, we devise two different strategies to train the discriminative
functions.

After scoring candidate arguments, the final global solution is built
by the inference module, which looks for the best scored argument
structure that satisfies the domain specific constraints.
Here, a global solution is a subset of candidate arguments, and its
score is defined as the sum of confidence values of the arguments that
form it.
We currently consider three constraints to determine which solutions
are valid:
\begin{enumerate}
\item[(a)] 
Candidate arguments for the same predicate can not overlap nor embed.
\vspace{-6pt}
\item[(b)] In a predicate, no duplicate arguments are allowed for the
numbered arguments {\tt A0-5}.
\vspace{-18pt}
\item[(c)] Arguments of a predicate can be embedded within
  arguments of other predicates but they can not overlap.
\end{enumerate}
The set of constraints can be extended with any other
rules, but in our particular case, we know that some constraints, e.g.,
providing only arguments indicated in the corresponding PropBank
frame, are already guaranteed by the individual models, and others,
e.g., constraints 3 and 4 in the previous sub-section,
have no positive impact on the overall performance (see
Section~\ref{sec:perfcs} for the empirical analysis). 
The inference algorithm we use is a bottom-up CKY-based dynamic programming
strategy~\cite{you67}. It builds the solution that maximizes the sum
of argument confidences while satisfying the constraints, in cubic
time. 

Next, we describe two different strategies to train the functions that
score candidate arguments. The first is a {\em local} strategy: each
function is trained as a binary batch classifier, independently of the
combination process which enforces the domain constraints. 
The second is a {\em global} strategy: functions are
trained as online rankers, taking into account the interactions that take
place during the combination process to decide between one argument or
another.

In both training strategies, the discriminative functions employ the same
representation of arguments, using
the complete feature set described in
Section~\ref{sec:features} (we analyze the contribution of each
feature group in Section~\ref{sec:results}). Our intuition was that
the rich feature space introduced in Section~\ref{sec:features} should
allow the gathering of sufficient statistics for robust scoring of the
candidate arguments. For example, the scoring classifiers might learn
that a candidate is to be trusted if: (a) two individual systems
proposed it, (b) if its label is {\tt A2} and it was generated by
Model~1, or (c) if it was proposed by Model~2 within a certain
argument sequence.

\subsubsection{Learning Local Classifiers}
\label{sec:modelloccl}

This combination process follows a cascaded architecture, in which
the learning component is decoupled from the inference module.
In particular, the training strategy consists of training a binary
classifier for each role label. The target of each label-based
classifier is to determine whether a candidate argument actually
belongs to the correct proposition of the corresponding predicate,
and to output a confidence value for this decision.

The specific training strategy is as follows. The training data
consists of a pool of labeled candidate arguments (proposed by
individual systems). Each candidate is either positive, in that it is
actually a correct argument of some sentence, or negative, if it is
not correct.  The strategy trains a binary classifier for each role
label $l$, independently of other labels. To do so, it concentrates on
the candidate arguments of the data that have label $l$. This forms a
dataset for binary classification, specific to the label $l$. With it,
a binary classifier can be trained using any of the existing
techniques for binary classification, with the only requirement that
our combination strategy needs confidence values with each binary
prediction. In Section~\ref{sec:results} we provide experiments using SVMs to
train such local classifiers.

In all, each classifier is trained independently of other classifiers
and the inference module. Looking globally at the combination process,
each classifier can be seen as an argument filtering component that
decides which candidates are actual arguments using a much richer
representation than the individual models. In this context, the
inference engine is used as a conflict resolution engine, to ensure
that the combined solutions are valid argument structures for
sentences.

\subsubsection{Learning Global Rankers}
\label{sec:modelgrank}

\newcommand{\Lab}{\mathcal{W}}
\newcommand{\A}{\mathcal{A}}
\newcommand{\w}{\hbox{\rm\bf w}}

This combination process couples learning and inference, i.e., the
scoring functions are trained to behave accurately within the
inference module.  In other words, the training strategy here is
global: the target is to train a global function that maps a set of
argument candidates for a sentence into a valid argument structure. In
our setting, the global function is a composition of scoring functions
 $-$one for each label, same as the previous strategy. 
Unlike the previous strategy, which is completely decoupled from the
inference engine, here the policy to map a set of candidates into a 
solution is that determined by the inference engine.

In recent years, research has been very active in global learning
methods for tagging, parsing and, in general, structure prediction
problems \cite{col02c,taskar03,taskar04,tsochan04}. 
In this article, we make use of the simplest technique for global
learning: an online learning approach that uses Perceptron
\cite{col02c}.
The general idea of the algorithm is similar to the original
Perceptron \cite{rosen58}: correcting the mistakes of a linear
predictor made while visiting training examples, in an additive
manner. The key point for learning global rankers relies on the
criteria that determines what is a mistake for the function being
trained, an idea that has been exploited in a similar way in
multiclass and ranking scenarios by \citeA{cramSi03b,cramSi03a}.

The Perceptron algorithm in our combination system works as follows 
(pseudocode of the algorithm is given in Figure~\ref{fig:gperc}).
Let $1 \ldots L$ be the possible role labels, and let 
\mbox{$\Lab$ $= \{ \w_1 \ldots \w_L \}$}
be the set of parameter vectors of the scoring
functions, one for each label.
Perceptron initializes the vectors in $\Lab$ to zero, and then
proceeds to cycle through the training examples, visiting one at a
time.
In our case, a training example is a pair $(y, \A)$, where $y$ is the
correct solution of the example and $\A$ is the set of candidate
arguments for it. Note that both $y$ and $\A$ are sets of labeled
arguments, and thus we can make use of the set difference. We
will note as $a$ a particular argument, as $l$ the label of $a$, and
as $\phi(a)$ the vector of features described in Section~\ref{sec:features}.
With each example, Perceptron performs two steps.  First, it predicts
the optimal solution $\hat{y}$ according to the current setting of
$\Lab$. Note that the prediction strategy employs the complete 
combination model, including the inference component.
Second, Perceptron corrects the vectors in $\Lab$ according to the
mistakes seen in $\hat{y}$: arguments with label $l$ seen in $y$ and
not in $\hat{y}$ are promoted in vector $\w_l$; on the other hand,
arguments in $\hat{y}$ and not in $y$ are demoted in $\w_l$.
This correction rule moves the scoring vectors towards missing
arguments, and away from predicted arguments that are not correct. It
is guaranteed that, as Perceptron visits more and more examples, this
feedback rule will improve the accuracy of the global combination
function when the feature space is almost linearly
separable~\cite{frSch99,col02c}. 

\begin{figure}[t]
\centering
\begin{tabular}{clc}
\hline
~~~~~ & \hspace*{0cm} {\bf Initialization:} {\it for each} $\w_l \in \Lab$ {\it do} $ \w_l = \mathbf{0}$ & ~~~~~\\
 & \hspace*{0cm} {\bf Training :} & \\
 & \hspace*{0.5cm} {\it for} $t = 1 \ldots T$ {\it do} & \\
 & \hspace*{1.0cm} {\it for each} training example $(y,  \A)$ {\it do} & \\
 & \hspace*{1.5cm} $\hat{y} = \mathrm{Inference}(\A,\Lab)$ & \\
% & \hspace*{1.0cm} Update: & \\
 & \hspace*{1.5cm} {\it for each} $a \in y\setminus\hat{y}$ {\it do} & \\
 & \hspace*{2.0cm}   let $l$ be the label of $a$ & \\
 & \hspace*{2.0cm}   $\w_l = \w_l + \phi(a)$ & \\
 & \hspace*{1.5cm} {\it for each} $a \in \hat{y}\setminus y$ {\it do} & \\
 & \hspace*{2.0cm}   let $l$ be the label of $a$ & \\
 & \hspace*{2.0cm}   $\w_l = \w_l - \phi(a)$ & \\
 & \hspace*{0cm} {\bf Output:} $\Lab$ & \\
\hline
\end{tabular}
\caption{Perceptron Global Learning Algorithm}
\label{fig:gperc}
\end{figure}

In all, this training strategy is global because the mistakes that
Perceptron corrects are those that arise when comparing the predicted structure
with the correct one. 
In contrast, a local strategy identifies mistakes looking individually
at the sign of scoring predictions: if some candidate argument is
(is not) in the correct solution and the current scorers predict a
negative (positive) confidence value, then the corresponding scorer is
corrected with that candidate argument. Note that this is the same
criteria used to generate training data for classifiers trained
locally. In Section~\ref{sec:results} we compare these
approaches empirically.

As a final note, for simplicity we have described Perceptron in its
most simple form. However, the Perceptron version we use in the
experiments reported in Section~\ref{sec:results} incorporates two
well-known extensions: kernels and averaging \cite{frSch99,col02a}. 
Similar to
SVM, Perceptron is a kernel method. That is, it can be represented in
dual form, and the dot product between example vectors can be
generalized by a kernel function 
that exploits richer representations.
On the other hand, averaging is a technique 
that increases the robustness of predictions during testing.
In the original form, test predictions are
computed with the parameters that result from the training process.
In the averaged version, test predictions are computed with an average
of all parameter vectors that are generated during training, after
every update.
Details of the technique can be found in the original article
of Freund \& Schapire.

\section{Experimental Results}
\label{sec:results}

In this section we analyze the performance of the three combination
strategies previously described: (a) inference with constraint
satisfaction, (b) learning-based inference with local rankers, and (c)
learning-based inference with global rankers. For the bulk of the
experiments we use candidate arguments generated by the three
individual SRL models described in Section~\ref{sec:base} and
evaluated in Section~\ref{sec:indperf}. 

\subsection{Experimental Settings}
\label{sec:expsettings}

All combination strategies (with one exception, detailed below) were
trained using the complete CoNLL-2005 training set (PropBank/Treebank
sections 2 to 21). To minimize the overfitting of the individual SRL
models on the training data, we partitioned the training corpus into
five folds and for each fold we used the output of the individual
models when trained on the remaining four folds. The models were tuned
on a separate development partition (Treebank section 24) and
evaluated on two corpora: (a) Treebank section 23, and (b) on the
three annotated sections of the Brown corpus. 

%%%%%%%%%%%%%% Rejection curves
\psfigyy{fig:rcurves}{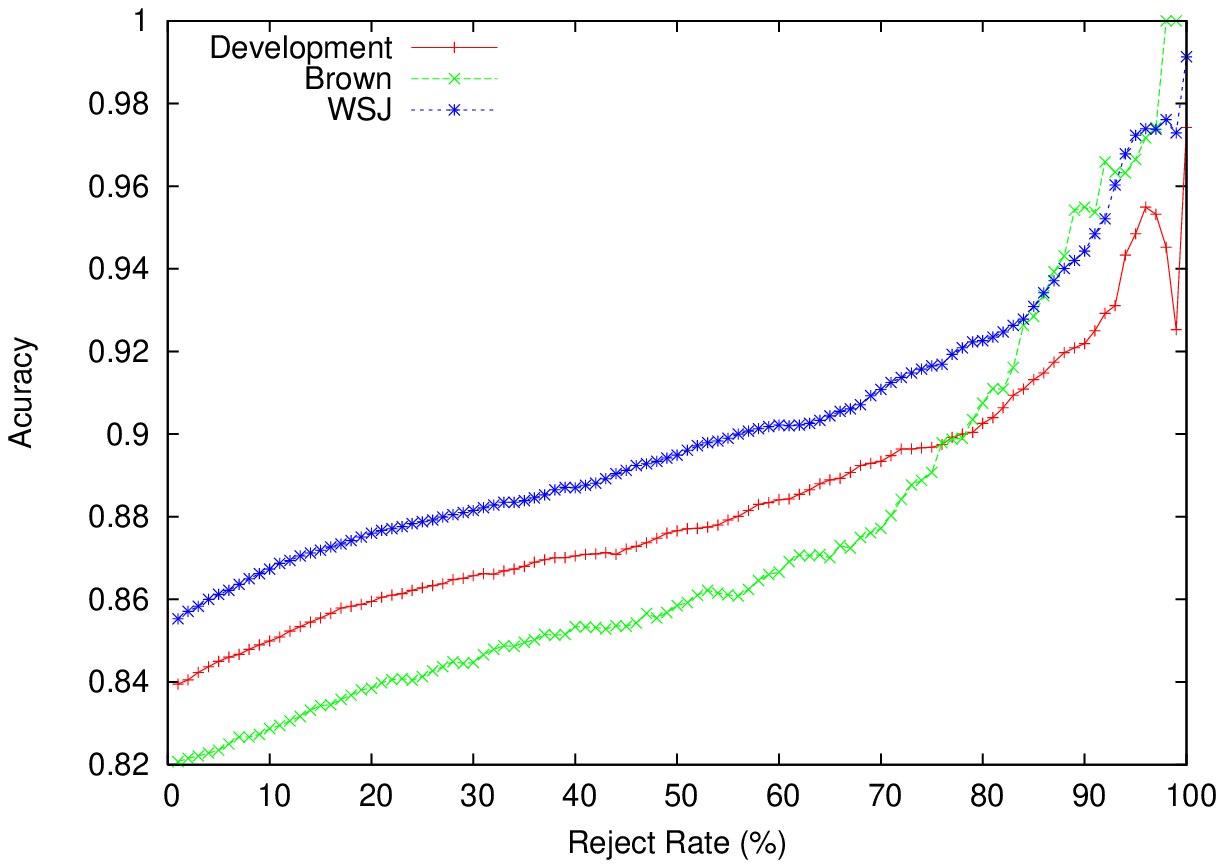}{0.2}{Rejection curves of the
  estimated output probabilities of the individual models.}{t}{3.5}

For the constraint satisfaction model,
we converted the scores of the output arguments of the three SRL
models into probabilities using the {\em softmax} function explained in
Section~\ref{sec:modelcs}. The development set (section 24) was used
to tune the $\gamma$ parameter of the {\em softmax} formula to a final value
of 0.1 for all models. In order to assess the quality of this
procedure, we plot in Figure~\ref{fig:rcurves} the rejection curves of
the estimated output probabilities with respect to classification
accuracy on the development and test sets (WSJ and Brown). To
calculate these plots, the probability estimates of all three models
are put together in a set and sorted in decreasing order. 
At a certain level of rejection ($n\%$),
the curve in Figure~\ref{fig:rcurves} plots the percentage of correct
arguments when the lowest scoring $n\%$ subset is rejected.
With few exceptions, the curves are increasing and
smooth, indicating a good correlation between probability estimates
and classification accuracy.

As a last experiment, in Section~\ref{sec:perflarger} we analyze the
behavior of the proposed combination strategies when the candidate
pool is significantly larger. For this experiment we used the top 10
best performing systems at the CoNLL-2005 shared task evaluation. 
In this setup there are two significant differences from the
experiments that used our in-house individual systems: (a) we had
access only to the systems' outputs on the PropBank development
section and on the two test sections, and (b) the argument probabilities of
the individual models were not available.
Thus, instead of the usual training set, we had to
train our combination models on the PropBank development section with
a smaller feature set. Note also
that the development set is only 3.45\% of the size of the regular
training set. We evaluated the resulting combination models on the
same two testing sections: WSJ and Brown.  

\subsection{Lower and Upper Bounds of the Combination Strategies}
\label{sec:bounds}

Before we venture into the evaluation of the combination strategies,
we explore the lower and upper bounds of the combinations models on
the given corpus and individual models. This analysis is important in
order to understand the potential of the proposed approach and to see
how close we actually are to realizing it. 

The performance upper bound is calculated with an oracle
combination system with a perfect filtering classifier that selects
only correct candidate arguments and discards all others.  
For comparison purposes, we have implemented a second oracle system
that simulates a {\em re-ranking} approach: for each predicate it
selects the candidate frame $-$i.e., the complete set of arguments for
the corresponding predicate proposed by a single model$-$ with the
highest F$_1$ score. 
Table~\ref{tab:upper} lists the results obtained on the WSJ and Brown
corpora by these two oracle systems using all three individual
models. 
The ``combination'' system is the oracle that simulates the
combination strategies proposed in this paper, which break
candidate frames and work with individual candidate arguments.  
Note that the precision of this oracle combination system is not 100\%
because in the case of discontinuous arguments, fragments that pass
the oracle filter are considered incorrect by the scorer when the
corresponding argument is not complete, e.g., an argument {\tt A1}
appears without the continuation {\tt C-A1}. 
The ``re-ranking'' columns list the results of the second oracle
system, which selects entire candidate frames. 

Table~\ref{tab:upper} indicates that the upper limit of the
combination approaches proposed in this paper
is relatively high: the F$_1$ of the combination oracle
system is over 14 points higher than our best individual system in the
WSJ test set, and over 17 points higher in the Brown corpus (see
Table~\ref{tab:base}).  
Furthermore, our analysis indicates that the potential of our
combination strategy is higher than that of re-ranking strategies,
which 
are limited to the performance of the best {\em complete} frame in
the candidate pool. By allowing the re-combination of arguments from
the individual candidate solutions this threshold is raised
significantly: over 6 F$_1$ points in WSJ and over 9 F$_1$ points in Brown.

Table~\ref{tab:contribwsj} lists the distribution of the
candidate arguments from the individual
models in the selection performed by the combination oracle system.
For conciseness, we list
only the core numbered arguments and we focus on the WSJ corpus. 
``$\cap$ of 3'' indicates the
percentage of correct arguments where all 3 models agreed, ``$\cap$ of
2'' indicates the percentage of correct arguments where any 2 models
agreed, and the other columns indicate the percentage of correct
arguments detected by a single model. Table~\ref{tab:contribwsj}
indicates that, as expected, two or more individual models agreed on a
large percentage of the correct arguments. Nevertheless, a significant
number of correct arguments, e.g., over 22\% of {\tt A3}, come from a
{\em single} individual system. This proves that, in order to achieve
maximum performance, one has to look beyond simple voting strategies
that favor arguments with high agreement between individual systems.

\begin{table}[t]
\centering
\begin{small}
\begin{tabular}{|l|c|c|c|c||c|c|c|c|}\cline{2-9}
\multicolumn{1}{l|}{} & \multicolumn{4}{|c||}{Combination} & \multicolumn{4}{|c|}{Re-Ranking} \\
\cline{2-9}
\multicolumn{1}{l|}{} & PProps & Precision & Recall & F$_1$ & PProps & Precision & Recall & F$_1$ \\
\hline
WSJ  & 70.76\% & 99.12\% & 85.22\% & 91.64 & 63.51\% & 88.08\% & 82.84\% & 85.38\\
Brown & 51.87\% & 99.63\% & 74.32\% & 85.14 & 45.02\% & 80.80\% & 71.70\% & 75.98\\
\hline
\end{tabular}
\end{small}
\caption{Performance upper limits detected by the two oracle systems.}
\label{tab:upper}
\end{table}

\begin{table}[t]
\centering
\begin{small}
\begin{tabular}{|l|c|c|c|c|c|}\cline{2-6}
\multicolumn{1}{l|}{} & $\cap$ of 3 & $\cap$ of 2 & Model 1 & Model 2 & Model 3 \\
\hline
{\tt A0}& 80.45\% & 12.10\% & 3.47\% & 2.14\% & 1.84\% \\
{\tt A1}& 69.82\% & 17.83\% & 7.45\% & 2.77\% & 2.13\% \\
{\tt A2}& 56.04\% & 22.32\% & 12.20\% & 4.95\% & 4.49\% \\
{\tt A3}& 56.03\% & 21.55\% & 12.93\% & 5.17\% &  4.31\% \\
{\tt A4}& 65.85\% & 20.73\% & 6.10\% & 2.44\% & 4.88\% \\
\hline
\end{tabular}
\end{small}
\caption{Distribution of the individual systems' arguments in the upper limit selection, for {\tt A0}$-${\tt A4} in the WSJ test set.}
\label{tab:contribwsj}
\end{table}

We propose two lower bounds for the performance of the combination models using two baseline systems:
\begin{itemize}
\itemsep -1mm
\item The first baseline is {\em recall-oriented}: it merges {\em all}
  the arguments generated by the individual systems. For conflict
  resolution, the baseline uses an approximate inference algorithm 
  consisting of
  two steps: (i) candidate arguments are sorted using a radix sort
  that orders the candidate arguments in descending order of: (a)
  number of models that agreed on this argument, (b) argument length
  in tokens, and (c) performance of the individual
  system\footnote{This combination produced the highest-scoring
    baseline model.}; (ii) Candidates are iteratively appended to the
  global solution only if they do not violate any of the domain
  constraints with the arguments already selected. 
\item The second baseline is {\em precision-oriented}: it considers
  only arguments where all three individual systems agreed. For
  conflict resolution it uses the same strategy as the previous
  baseline system. 
\end{itemize}
Table~\ref{tab:lower} shows the performance of these two baseline
models. As expected, the precision-oriented baseline obtains a
precision significantly higher than the best individual model
(Table~\ref{tab:base}), but its recall suffers because the individual
models do not agree on a fairly large number of candidate
arguments. The recall-oriented baseline is more balanced: as expected
the recall is higher than any individual model and the precision does
not drop too much because the inference strategy filters out many
unlikely candidates. Overall, the recall-oriented baseline performs
best, with an F$_1$ 1.22 points higher than the best individual model
on the WSJ corpus, and 0.41 points lower on the Brown corpus. 

\begin{table}[t]
%\begin{small}
\begin{center}
\begin{tabular}{|l|c|c|c|c|}\cline{2-5}
\multicolumn{1}{l|}{WSJ} & PProps & Prec. & Recall & F$_1$\\
\hline
baseline recall & {\bf 53.71}\% & 78.09\% & {\bf 78.77}\% & 78.43 {\scriptsize$\pm 0.8$}\\
baseline precision & 35.43\% & {\bf 92.49}\% & 60.48\% & 73.14 {\scriptsize$\pm 0.9$}\\
\hline
%\end{tabular}
%\begin{tabular}{|l|c|c|c|c|}\cline{2-5}
\multicolumn{5}{l}{Brown}\\% & PProps & Prec. & Recall & F$_1$\\
\hline
baseline recall & {\bf 36.94}\% & 68.57\% & {\bf 66.05}\% & 67.29 {\scriptsize$\pm 2.0$} \\
baseline precision & 20.52\% & {\bf 88.74}\% & 46.35\% & 60.89 {\scriptsize$\pm 2.1$}\\
\hline
\end{tabular}
%\end{small}
\end{center}
\caption{Performance of the baseline models on the WSJ and Brown test sets.}
\label{tab:lower}
\end{table}

\subsection{Performance of the Combination System with Constraint Satisfaction}
\label{sec:perfcs}

In the Constraint Satisfaction setting the arguments output by
individual Models 1, 2, and 3 are recombined into an expected
better solution that satisfies
a set of constraints. 
We have run the inference model based on Constraint Satisfaction 
described in Section~\ref{sec:modelcs} using the Xpress-MP ILP 
solver\footnote{Xpress-MP is a Dash Optimization product that is 
free for academic usage.}. 
The main results are summarized in Table~\ref{tab:localglobal}.
The variants presented in that table
are the following: ``Pred-by-pred'' stands for local
optimization, which processes each verb predicate independently
from others, while ``Full sentence'' stands for global
optimization, i.e., resolving all the verb predicates of the
sentence at the same time. The column labeled ``Constraints''
shows the particular constraints applied at each
configuration. The ``$O$'' column presents the value of the
parameter for correcting the bias towards candidate
overgeneration. Concrete values are empirically set to maximize
the F$_1$ measure on the development set. $O=0$ corresponds to a
setting in which no bias correction is applied.

\begin{table}[t]
\centering
\begin{tabular}{|l|l|l|c|c|c|c|}\cline{2-7}
\multicolumn{1}{l|}{WSJ} & Constraints & $O$ &PProps & Precision & Recall & F$_1$\\
\hline
Pred-by-pred  & 1      & 0.30 & 52.29\% & 84.20\% & 75.64\% & 79.69 {\scriptsize$\pm 0.8$} \\
              & 1+2    & 0.30 & 52.52\% & 84.61\% & 75.53\% & 79.81 {\scriptsize$\pm 0.6$} \\
              & 1+2+3  & 0.25 & 52.31\% & 84.34\% & 75.48\% & 79.67 {\scriptsize$\pm 0.7$} \\
              & 1+2+4  & 0.30 & 51.40\% & 84.13\% & 75.04\% & 79.32 {\scriptsize$\pm 0.8$} \\
              & 1+2+3+4& 0.30 & 51.19\% & 83.86\% & 74.99\% & 79.18 {\scriptsize$\pm 0.7$} \\\hline
Full sentence & 1+2+5  & 0.30 & 52.53\% & 84.63\% & 73.53\% & {\bf 79.82} {\scriptsize$\pm 0.7$} \\
              & 1+2+6  & 0.30 & 52.48\% & 84.64\% & 75.51\% & 79.81 {\scriptsize$\pm 0.8$} \\
              & 1+2+5+6& 0.30 & 52.50\% & {\bf 84.65}\% & 75.51\% & {\bf 79.82} {\scriptsize$\pm 0.6$} \\
              & 1+2+5+6& 0    & {\bf 54.49}\% & 78.74\% & {\bf 79.78}\% & 79.26 {\scriptsize$\pm 0.7$} \\\hline
\multicolumn{7}{l}{Brown}\\\hline
Full sentence & 1+2+5+6& 0.30 & 35.70\% & {\bf 78.18}\% & 62.06\% & {\bf 69.19} {\scriptsize$\pm 2.1$} \\
              & 1+2+5+6& 0    & {\bf 38.06}\% & 69.80\% & {\bf 67.85}\% & 68.81 {\scriptsize$\pm 2.2$} \\
\hline
\end{tabular}
\caption{Results, on WSJ and Brown test sets, obtained by multiple
  variants of the constraint satisfaction approach}
\label{tab:localglobal}
\end{table}

Some clear conclusions can be drawn from
Table~\ref{tab:localglobal}. First, we observe that any
optimization variant obtains F$_1$ results above 
both
the individual
systems (Table~\ref{tab:base}) and the baseline combination
schemes (Table~\ref{tab:lower}). 
The best combination model scores 2.61 F$_1$ points
in WSJ and 1.49 in Brown higher than the best individual system.
Taking into account that no learning is
performed, it is clear that Constraint Satisfaction is a simple
yet formal setting that achieves good results.

A somewhat surprising result is that all performance improvements
come from constraints 1 and 2 (i.e., no overlapping nor embedding
among arguments of the same verb, and no repetition of core
arguments in the same verb). Constraints 3 and 4 are
harmful, while the sentence-level constraints (5 and 6) 
have no impact on the overall performance\footnote{In
Section~\ref{sec:perfwfeed} we will see that when the learning
strategy incorporates global feedback, performing a sentence-level
inference is slightly better than proceeding predicate by
predicate.}. 
Our analysis of the proposed constraints yielded the following 
explanations:
 
\begin{itemize} 

\item Constraint number 3 prevents 
  the assignment of an {\tt R-X} argument 
  when the referred argument {\tt X} is not present. This makes the
  inference miss some easy {\tt R-X} arguments when the {\tt X} argument
  is not correctly identified (e.g., constituents that start with
  \{{\em that, which, who}\} followed by a verb are always {\tt R-A0}). 
  Furthermore, this
  constraint presents a lot of exceptions: up to $18.75\%$ of the
  {\tt R-X} arguments in the WSJ test set do not have the referred
  argument {\tt X} (e.g., ``{\em when}'' in ``{\em the law tells them
  when to do so}''), therefore the hard application of constraint
  3 prevents the selection of some correct {\tt R-X} candidates.  The
  official evaluation script from CoNLL-2005 ({\em srl-eval})
  does not require this constraint to be satisfied to consider a
  solution consistent.

\item The {\em srl-eval} script requires that constraint number 4 (i.e., a
  {\tt C-X} tag is not accepted without a preceding {\tt X} argument) be fulfilled 
  for a candidate solution to be considered consistent. 
  But when it  finds
  a solution violating the constraint its behavior is to convert
  the first {\tt C-X} (without a preceding {\tt X}) into {\tt X}. It turns out
  that this simple post-processing strategy is better than
  forcing a coherent solution in the inference step
  because it
  allows to recover from the error when an argument has been
  completely recognized but labeled only with {\tt C-X} tags. 

\item Regarding sentence-level constraints, we observed that in our
 setting, inference using local constraints (1+2) rarely produces a
 solution with inconsistencies at sentence level.\footnote{This fact
   is partly explained by the small number of overlapping arguments
   in the candidate pool produced by the three individual models.}
 This makes constraint 5 useless since it is almost never violated.
 Constraint number 6 (i.e., no sharing of {\tt AM}s among different
 verbs) is more ad-hoc and represents a less universal principle in
 SRL.  The number of exceptions to that constraint, in the WSJ test
 set, is $3.0\%$ for the gold-standard data and $4.8\%$ in the output
 of the inference that uses only local constraints (1+2). Forcing
 the fulfillment of this constraint makes the inference process
 commit as many errors as corrections, making its effect negligible.

\end{itemize}

Considering that some of the constraints are not universal, 
i.e., exceptions exist in the gold standard, 
it seems reasonable to convert them
into \emph{soft} constraints. This can be done by precomputing their
compatibility from corpora counts using, for instance, point-wise
mutual information, and incorporating its effect in the compatibility
function as explained in section \ref{sec:modelcs}. This softening
could, in principle, increase the overall recall of the combination.
Unfortunately, our initial experiments showed no differences between
the hard and soft variants.

Finally, the differences between the optimized values of the bias
correcting parameter and $O=0$ are clearly explained by observing
precision and recall values. The default version tends to overgenerate
argument assignments, which implies a higher recall at a cost of a
lower precision. On the contrary, the F$_1$ optimized variant is more
conservative and needs more evidence to select a candidate. As a
result, the precision is higher but the recall is lower. A side effect
of being restrictive with argument assignments, is that the number of
correctly annotated complete propositions is also lower in the
optimized setting.

%%%%%%%%%%%%%% Precision-Recall curves
\begin{figure}[t]
\begin{center}
\epsfxsize = 2.9 in \epsfbox{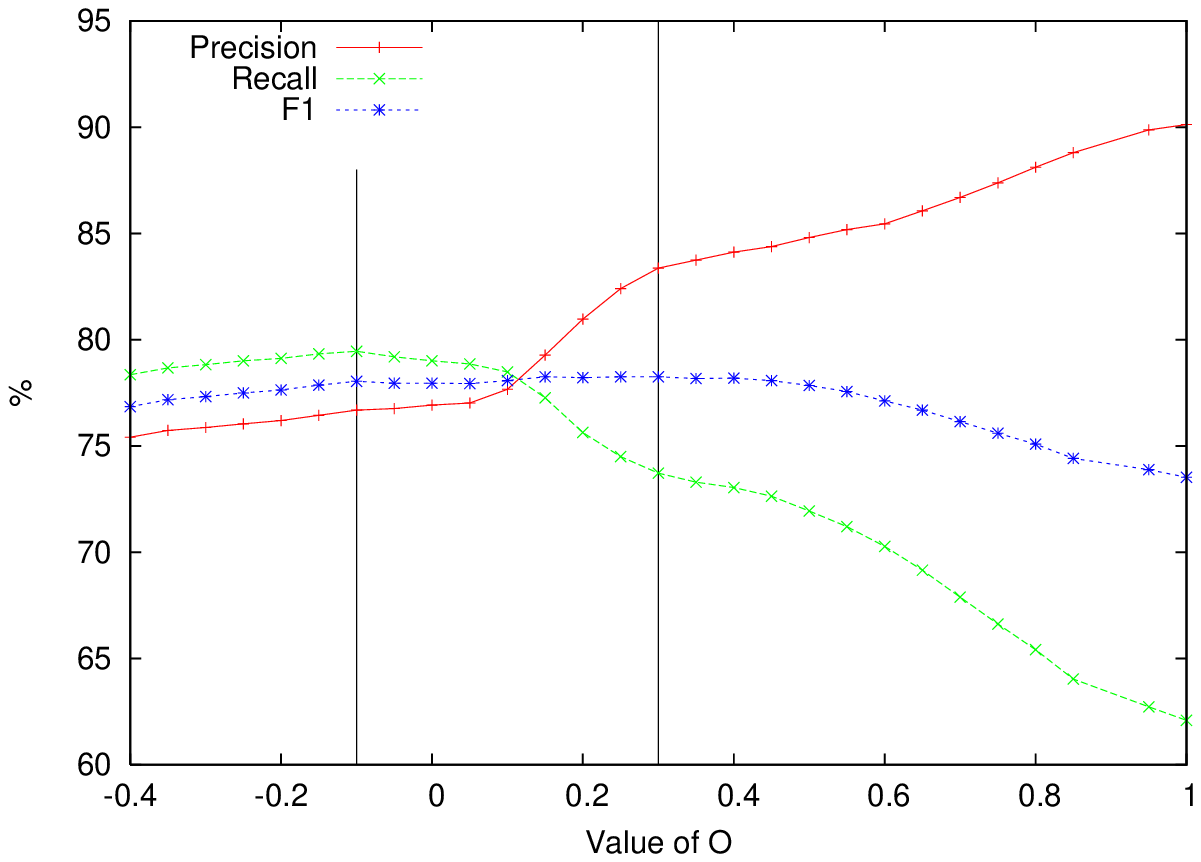}
\epsfxsize = 2.9 in \epsfbox{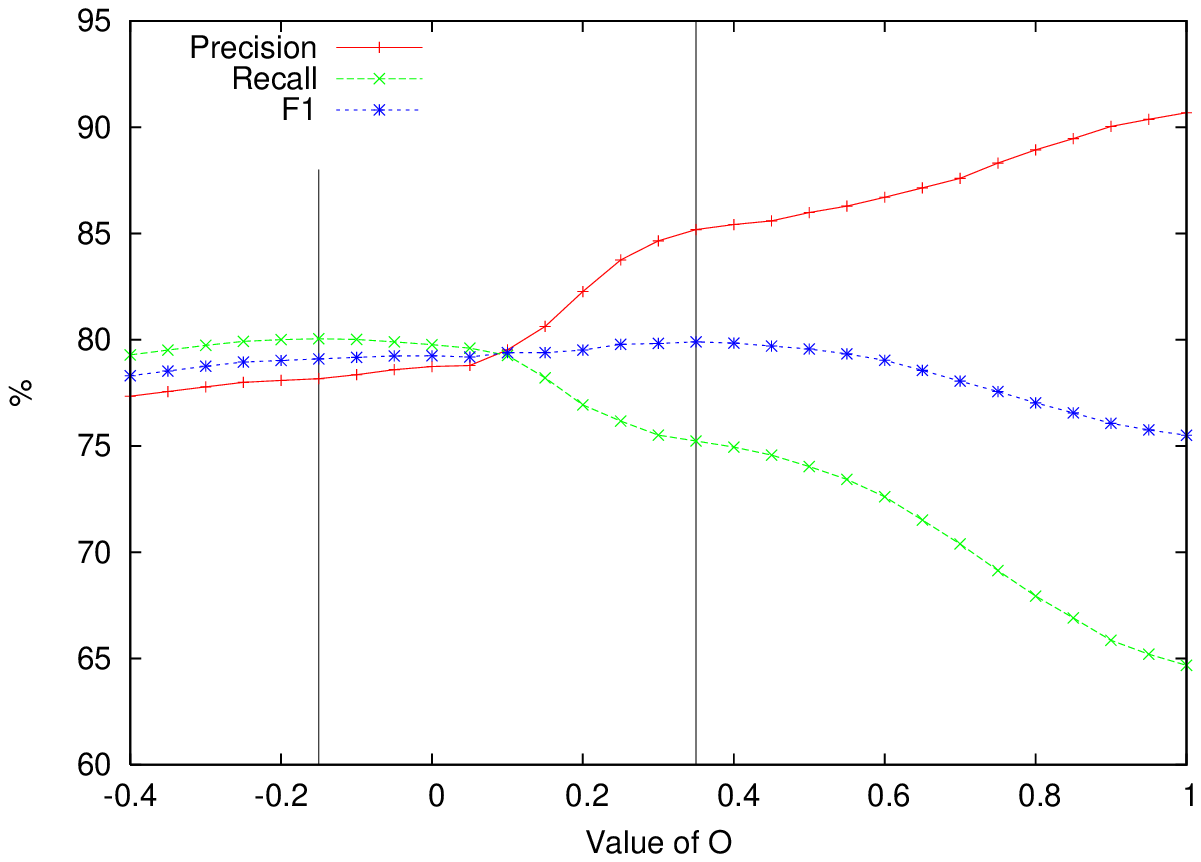} 
\end{center}
\vspace{0.1in}
\caption{Precision-Recall plots, with respect to the bias correcting parameter ($O$), 
for the WSJ development and test sets (left and right plots, respectively).}
\label{fig:prplots}
\end{figure}

The preference for a high-precision vs. a high-recall system is mostly
task-dependant. It is interesting to note that in this constraint
satisfaction setting, adjusting the precision--recall tradeoff can be
easily done by varying the value of the bias correcting score. In
Figure~\ref{fig:prplots}, we plot the precision--recall curves with
respect to different values of the $O$ parameter (the optimization is
done using constraints 1, 2, 5, and 6).
As expected, high values of $O$ promote precision and demote recall,
while lower values of $O$ do just the contrary.  Also, we see that
there is a wide range of values for which the combined F$_1$ measure
is almost constant (the approximate intervals are marked using
vertical lines), making it possible to select different recall and
precision values with a global performance (F$_1$) near to the
optimal.  Parenthetically, note also that the optimal value estimated
on the development set ($O=0.3$) generalizes very well to the WSJ test
set.

%%%%%%%%%%%%%%%%%%%%%%%%%%%%%%%%%%%%%%%%%%%%%%%%%%

\subsection{Performance of the Combination System with Local Rankers}
\label{sec:perfwofeed}

We implemented the candidate-scoring classifiers for this combination
strategy using Support Vector Machines (SVM) with polynomial kernels
of degree 2, which performed slightly better than other types of SVMs
or AdaBoost. 
We have implemented the SVM classifiers with the SVM$^{light}$ software\footnote{\url{http://svmlight.joachims.org/}}. Outside of changing the default kernel to polynomial we have not modified the default parameters.
For the experiments reported in this section, we trained
models for all 4 possible combinations of our 3 individual systems,
using the complete feature set introduced in
Section~\ref{sec:features}. The dynamic programming engine used for
the actual inference processes each predicate independently (similar
to the ``Pred-by-pred'' approach in the previous sub-section).

Table~\ref{tab:combo} summarizes the performance of the combined
systems on the WSJ and Brown corpora.
%\footnote{For conciseness, in Table~\ref{tab:combo} we introduced the notation M1+2+3 to indicate the combination of Models 1, 2, and 3.} 
Table~\ref{tab:combo} indicates that our combination strategy is always
successful: the results of all combination systems improve upon their
individual models (Table~\ref{tab:base})
and their F$_1$ scores are always better than 
the baselines' (Table~\ref{tab:lower}).
The last column in the table shows the F$_1$ improvement of the combination model w.r.t. the best individual model in each set.
As expected, the highest scoring combined
system includes all three individual models. Its F$_1$ measure
is 3.35 points higher than the best individual model (Model~2) in the
WSJ test set and 2.38 points higher in the Brown test set. 
Note that with any combination of two individual systems we outperform
the current state of the art (see Section~\ref{sec:related} for
details). This is empirical proof that robust and successful
combination strategies for the SRL problem are possible. 
Table~\ref{tab:combo} also indicates that, even though the partial parsing model (Model~1) is the worst performing individual model, its contribution to the ensemble is very important, indicating that the information it provides is indeed complementary to the other models'. For instance, in WSJ the performance of the combination of the two best individual models (Models~2+3) is worse than the combinations using model 1 (Models~1+2 and 1+3).

\begin{table}[t]
%\begin{small}
\begin{center}
\begin{tabular}{|l|c|c|c|c|c|}\cline{2-6}
\multicolumn{1}{l|}{WSJ} & PProps & Prec. & Recall & F$_1$ & F$_1$ improvement\\
\hline
Models 1+2  & 49.28\% & 87.39\% & 72.88\% & 79.48 {\scriptsize$\pm 0.6$} & +2.27\\
Models 1+3  & 48.26\% & 86.80\% & 73.20\% & 79.42 {\scriptsize$\pm 0.6$} & +2.96\\
Models 2+3  & 49.36\% & 86.63\% & 73.03\% & 79.25 {\scriptsize$\pm 0.7$} & +2.04\\
Models 1+2+3  & {\bf 51.66\%} & {\bf 87.47\%} & {\bf 74.67\%} & {\bf 80.56} {\scriptsize$\pm 0.6$} & +3.35\\
\hline
%\end{tabular}
%\begin{tabular}{|l|c|c|c|c|}\cline{2-5}
\multicolumn{6}{l}{Brown}\\% & PProps & Prec. & Recall & F$_1$\\
\hline
Models 1+2  & 34.33\% & 81.14\% & 60.86\% & 69.55 {\scriptsize$\pm 2.0$} & +1.85 \\
Models 1+3  & 31.22\% & 80.43\% & 59.07\% & 68.11 {\scriptsize$\pm 1.9$} & +2.69\\
Models 2+3  & 32.84\% & 80.90\% & 60.31\% & 69.11 {\scriptsize$\pm 2.1$} & +1.41\\
Models 1+2+3  & {\bf 34.33\%} & {\bf 81.75\%} & {\bf 61.32\%} & {\bf 70.08} {\scriptsize$\pm 2.1$} & +2.38\\
\hline
\end{tabular}
%\end{small}
\end{center}
\caption{Overall results of the learning-based inference with local rankers on the WSJ and Brown test sets.}
\label{tab:combo}
\end{table}

\begin{table}[t]
\centering
\begin{tabular}{|r|c|c|c|c|}\cline{2-5}
\cline{2-5}
\multicolumn{1}{l|}{WSJ} & PProps & Prec. & Recall & F$_1$ \\
\hline
FS1 & 50.24\% & 86.47\% & 73.51\% & 79.47 {\scriptsize$\pm 0.7$}\\
\hline
+ FS2 & 50.39\% & 86.41\% & 73.68\% & 79.54 {\scriptsize$\pm 0.6$}\\
\hline
+ FS3 & 51.22\% & 86.13\% & 74.35\% & 79.80 {\scriptsize$\pm 0.7$}\\
\hline
+ FS4 & 50.66\% & 86.67\% & 74.10\% & 79.89 {\scriptsize$\pm 0.7$}\\
\hline
+ FS5 & 51.38\% & 87.21\% & 74.61\% & 80.42 {\scriptsize$\pm 0.6$}\\
\hline
+ FS6 & {\bf 51.66\%} & {\bf 87.47\%} & {\bf 74.67\%} & {\bf 80.56} {\scriptsize$\pm 0.6$}\\
\hline
%\end{tabular}
%\begin{tabular}{|r|c|c|c|c|}\cline{2-5}
%\cline{2-5}
\multicolumn{5}{l}{Brown}\\% & PProps & Prec. & Recall & F$_1$ \\
\hline
FS1 & 32.21\% & 80.12\% & 59.44\% & 68.25 {\scriptsize$\pm 2.0$}\\
\hline
+ FS2 & 32.84\% & 80.80\% & 59.94\% & 68.83 {\scriptsize$\pm 2.2$}\\
\hline
+ FS3 & 33.33\% & 80.29\% & 60.82\% & 69.21 {\scriptsize$\pm 2.0$}\\
\hline
+ FS4 & 33.33\% & 81.10\% & 60.50\% & 69.30 {\scriptsize$\pm 2.1$}\\
\hline
+ FS5 & 34.08\% & 81.76\% & 61.14\% & 69.96 {\scriptsize$\pm 2.2$}\\
\hline
+ FS6 & {\bf 34.33\%} & {\bf 81.75\%} & {\bf 61.32\%} & {\bf 70.08} {\scriptsize$\pm 1.9$}\\
\hline

\end{tabular}
\caption{Feature analysis for the learning-based inference with local rankers.}
\label{tab:wofeed-featanal}
\end{table}

Due to its simple architecture $-$i.e., no feedback from the conflict
resolution component to candidate filtering$-$ this inference model is
a good framework to study the contribution of the features proposed in
Section~\ref{sec:features}. For this study we group the features into
6 sets: FS1 $-$the voting features, FS2 $-$the overlap features with
arguments for the same predicate, FS3 $-$the overlap features with
arguments for other predicates, FS4 $-$the partial-syntax features,
FS5 $-$the full-syntax features, and FS6 $-$the probabilities
generated by the individual systems for the candidate arguments. Using
these sets we constructed 6 combination models by increasing the
number of features made available to the argument filtering
classifiers, e.g., the first system uses only FS1, the second system
adds FS2 to the first system's features, FS3 is added for the third
system, etc. Table~\ref{tab:wofeed-featanal} lists the performance of
these 6 systems for the two test corpora. This empirical analysis
indicates that the feature sets with the highest contribution are: 
\begin{itemize}
\item FS1, which boosts the F$_1$ score of the combined system 2.26
  points (WSJ) and 0.55 points (Brown) over our best individual
  system. This is yet another empirical proof that voting is a
  successful combination strategy. 
\vspace{-0.07in}
\item FS5, with a contribution of 0.53 points (WSJ) and 0.66 points
  (Brown) to the F$_1$ score. These numbers indicate that the
  filtering classifier is capable of learning some of the ``preferences''
  of the individual models for certain syntactic structures.  
\vspace{-0.07in}
\item FS3, which contributes 0.26 points (WSJ) and 0.38 points (Brown)
  to the F$_1$ score. These results promote the idea that information
  about the overall sentence structure, in our case inter-predicate
  relations, can be successfully used for the problem of SRL. 
To our knowledge, this is novel.
\end{itemize}
All the proposed features have a positive contribution to the
performance of the combined system. Overall, we achieve an F$_1$ score
that is 1.12 points (WSJ) and 2.33 points (Brown) higher than the best
performing combined system at the CoNLL-2005 shared task evaluation
(see Section~\ref{sec:related} for details). 

\subsection{Performance of the Combination System with Global Rankers}
\label{sec:perfwfeed}

In this section we report experiments with the global Perceptron
algorithm described in Section~\ref{sec:modelgrank}, that globally
trains the scoring functions as rankers. Similar to the local SVM models, we
use polynomial kernels of degree 2. Furthermore, the predictions at
test time used averages of the parameter vectors, following the
technique of \citeA{frSch99}. 

We were interested in two main aspects. 
First, we evaluate the effect of training the scoring functions with Perceptron
using two different update rules, one global and the other local. The
global feedback rule, detailed in Section~\ref{sec:modelgrank},
corrects the mistakes found when comparing the correct argument
structure with the one that results from the inference (this is noted as
``global'' feedback). In contrast, the local feedback rule
corrects the mistakes found {\em before} inference, 
when each candidate argument is handled independently, ignoring the 
global argument structure generated
(this is noted as ``local'' feedback).
Second, we analyze the effect of using different constraints in the inference
module. To this extent, we configured the inference module in two
ways. The first processes the predicates of a sentence independently,
and thus 
might select overlapping arguments of different predicates, which is 
incorrect according to the domain constraints
(this one is noted as ``Pred-by-pred'' inference). The
second processes all predicates jointly, and enforces a hierarchical
structure of arguments, 
where arguments never overlap, and arguments of a predicate are
allowed to embed arguments of other predicates
(this is noted as ``Full sentence'' inference).
From this perspective, the model with local update and
``Pred-by-pred'' inference is almost identical to the local
combination strategy described in Section~\ref{sec:perfwofeed}, with
the unique difference that here we use Perceptron instead of SVM. This
apparently minute difference turns out to be significant for our
empirical analysis because it allows us to measure the contribution of
both SVM margin maximization and global feedback to the
classifier-based combination strategy (see
Section~\ref{sec:perfdisc}). 

We trained four different models: with local or global feedback, and
with predicate-by-predicate or joint inference.
Each model was trained for 5 epochs on the training data, and
evaluated on the development data after each training epoch.
We selected the best performing point on development, and evaluated
the models on the test data. Table~\ref{tab:perfwfeed} reports the
results on test data.

\begin{table}[t]
\centering
\begin{tabular}{|r|c|c|c|c|}\cline{2-5}
\cline{2-5}
\multicolumn{1}{l|}{WSJ} & PProps & Prec. & Recall & F$_1$ \\
\hline
Pred-by-pred, local    & 50.71\% &  {\bf 86.80\%} &  74.31\% &  80.07 {\scriptsize$\pm 0.7$}\\
Full sentence,   local    & 50.67\% &  {\bf 86.80\%} &  74.29\% &  80.06 {\scriptsize$\pm 0.7$}\\
Pred-by-pred, global   & 53.45\% &  84.66\% &  76.19\% &  80.20 {\scriptsize$\pm 0.7$}\\
Full sentence,   global   & {\bf 53.81\%} &  84.84\% &  {\bf 76.30\%} &  {\bf 80.34} {\scriptsize$\pm 0.6$}\\
\hline
%\end{tabular}
%\begin{tabular}{|r|c|c|c|c|}\cline{2-5}
%\cline{2-5}
\multicolumn{5}{l}{Brown}\\% & PProps & Prec. & Recall & F$_1$ \\
\hline
Pred-by-pred, local    & 33.33\% & 80.62\% & 60.77\% &  69.30 {\scriptsize$\pm 1.9$}\\
Full sentence,   local    & 33.33\% & {\bf 80.67\%} & 60.77\% &  69.32 {\scriptsize$\pm 2.0$}\\
Pred-by-pred, global   & 35.20\% &  77.65\% & 62.70\% &  69.38 {\scriptsize$\pm 1.9$}\\
Full sentence,   global   & {\bf 35.95\%} &  77.91\% & {\bf 63.02\%} &  {\bf 69.68} {\scriptsize$\pm 2.0$}\\
\hline
\end{tabular}

\caption{Test results of the combination system with global
  rankers. Four configurations are evaluated, that combine
  ``Pred-by-pred'' or ``Full sentence'' inference with ``local'' or
  ``global'' feedback. } 
\label{tab:perfwfeed}
\end{table}

Looking at results, a first impression is that the difference in
F$_1$ measure is not very significant among different
configurations. However, some observations can be pointed out.
Global methods
achieve much better recall figures, whereas local methods prioritize the
precision of the system. 
Overall, global methods achieve a more
balanced tradeoff between precision and recall, which contributes to a
better F$_1$ measure. 

Looking at ``Pred-by-pred'' versus ``Full sentence'' inference, it 
can be seen that only
the global methods are sensitive to the difference.
Note that a local model is trained independently of the inference
module. Thus, adding more constraints to the inference engine does not
change the parameters of the local model. At testing time, the
different inference configurations do not affect the results.
In contrast, the global models are trained dependently of the
inference module. When moving from ``Pred-by-pred'' to
``Full sentence'' inference, consistency is enforced between argument
structures of different predicates, and this benefits both the
precision and recall of the method.  The global learning algorithm
improves both in precision and recall when coupled with a joint
inference process that considers more constraints in the solution.

Nevertheless, the combination system with local SVM classifiers,
presented in the previous section, achieves marginally better
F$_1$ score than the global learning method (80.56\% vs.
80.34\% in WSJ). 
This is explained by the different machine learning algorithms 
(we discuss this issue in detail in Section~\ref{sec:perfdisc}). 
The better F$_1$ score is accomplished by a much better
precision in the local approach (87.47\% vs. 84.84\% in WSJ), whereas
the recall is lower in the local than in the global approach
(74.67\% vs. 76.30\% in WSJ). 
On the other hand, the global strategy produces more completely-correct
annotations (see the ``PProps'' column) than any of the local
strategies investigated (see Tables~\ref{tab:perfwfeed}
and~\ref{tab:combo}). 
This is to be expected, considering that the global strategy optimizes
a sentence-level cost function. Somewhat surprisingly, the number of
perfect propositions generated by the global strategy is lower than
the number of perfect propositions produced by the
constraint-satisfaction approach. We discuss this result in
Section~\ref{sec:perfdisc}. 

\subsection{Scalability of the Combination Strategies}
\label{sec:perflarger}

\begin{table}[t]
  \centering
\begin{small}
\begin{tabular}{|l|c|c|c|c|c||c|c|c|c|}\cline{3-10}
\multicolumn{2}{l|}{} & \multicolumn{4}{|c||}{WSJ} & \multicolumn{4}{|c|}{Brown}\\
\cline{3-10}
\multicolumn{2}{l|}{} & PProps & Prec. & Recall & F$_1$ & PProps & Prec. & Recall & F$_1$\\
\hline
koomen &$\surd$ & 53.79\% & 82.28\% & 76.78\% & {\bf 79.44} & 32.34\% & 73.38\% & 62.93\% & 67.75 \\
pradhan+ &$\surd$ & 52.61\% & {\bf 82.95}\% & 74.75\% & 78.63 & {\bf 38.93}\% & {\bf 74.49}\% & 63.30\% & {\bf 68.44} \\
haghighi &$\surd$ & {\bf 56.52}\% & 79.54\% & {\bf 77.39}\% & 78.45 & 37.06\% & 70.24\% & {\bf 65.37}\% & 67.71 \\
marquez &$\surd$ & 51.85\% & 79.55\% & 76.45\% & 77.97 & 36.44\% & 70.79\% & 64.35\% & 67.42 \\
{\footnotesize pradhan} & {\Large $\times$} & {\footnotesize  50.14\%} & {\footnotesize  81.97\%} & {\footnotesize  73.27\%} & {\footnotesize  77.37} & {\footnotesize  36.44\%} & {\footnotesize  73.73\%} & {\footnotesize  61.51\%} & {\footnotesize 67.07} \\
surdeanu &$\surd$ & 45.28\% & 80.32\% & 72.95\% & 76.46 & 29.48\% & 72.41\% & 59.67\% & 65.42 \\
tsai &$\surd$ & 45.43\% & 82.77\% & 70.90\% & 76.38 & 30.47\% & 73.21\% & 59.49\% & 65.64 \\
che &$\surd$ & 47.81\% & 80.48\% & 72.79\% & 76.44 & 31.84\% & 71.13\% & 59.99\% & 65.09 \\
moschitti &$\surd$ & 47.66\% & 76.55\% & 75.24\% & 75.89 & 30.85\% & 65.92\% & 61.83\% & 63.81 \\
tjongkimsang &$\surd$ & 45.85\% & 79.03\% & 72.03\% & 75.37 & 28.36\% & 70.45\% & 60.13\% & 64.88 \\
{\footnotesize yi} & {\Large $\times$} & {\footnotesize 47.50\%} & {\footnotesize 77.51\%} & {\footnotesize 72.97\%} & {\footnotesize 75.17} & {\footnotesize 31.09\%} & {\footnotesize 67.88\%} & {\footnotesize 59.03\%} & {\footnotesize 63.14} \\
ozgencil &$\surd$ & 46.19\% & 74.66\% & 74.21\% & 74.44 & 31.47\% & 65.52\% & 62.93\% & 64.20 \\
\hline
\end{tabular}
\end{small}
\caption{Performance of the best systems at CoNLL-2005. The
  ``pradhan+'' contains post-evaluation improvements. The top 5
  systems are actually combination models themselves. The second
  column marks with $\times$ the systems not used in our evaluation:
  ``pradhan'', which was replaced by its improved version
  ``pradhan+'', and ``yi'', due to format errors in the submitted
  data.} 
\label{tab:10best}
\end{table}

All the combination experiments reported up to this point used the
candidate arguments generated by the three individual SRL models
introduced in Section~\ref{sec:base}. While these experiments do
provide an empirical comparison of the three inference models
proposed, they do not answer an obvious scalability question: how do
the proposed combination approaches scale when the number of candidate
arguments increases but their quality diminishes? We are mainly
interested in answering this question for the last two combination
models (which use inference based on learning with local or global
rankers) for two reasons: (a) they performed better than the
constraint satisfaction model in our previous experiments, and (b)
because they have no requirements on the individual SRL systems'
outputs $-$unlike the constraint satisfaction model which requires
the argument probabilities of the individual models$-$ they can be coupled
to pools of candidates generated by any individual SRL model.

For this scalability analysis, we use as individual SRL models the top
10 systems at the CoNLL-2005 shared task
evaluation. Table~\ref{tab:10best} summarizes the performance of these
systems on the same two test corpora used in our previous
experiments. As Table~\ref{tab:10best} indicates, the performance of
these systems varies widely: there is a difference of 5 F$_1$ points
in the WSJ corpus and of over 4 F$_1$ points in the Brown corpus
between the best and the worst system in this set.  

For the combination experiments we generated 5 candidate pools using
the top 2, 4, 6, 8, and 10 individual systems labeled with $\surd$ in
Table~\ref{tab:10best}. We had to make two changes to the experimental
setup used in the first part of this section: (a) we trained our
combined models on the PropBank development section because we did not
have access to the individual systems' outputs on the PropBank
training partition; and (b) from the feature set introduced in
Section~\ref{sec:features} we did not use the individual systems'
argument probabilities because the raw activations of the individual
models' classifiers were not available. Note that under these settings the size of the training corpus is 10 times smaller than the size of the training set used in the previous experiments.

Table~\ref{tab:oracletop10} shows the upper limits of these setups
using the ``combination'' and ``re-ranking'' oracle systems introduced
in Section~\ref{sec:bounds}. 
Besides performance numbers, we also list in
Table~\ref{tab:oracletop10} the average number of candidates per
sentence for each setup, i.e., number of unique candidate arguments
(``\#~Args./Sent.'') for the ``combination'' oracle and number of
unique candidate frames (``\#~Frames/Sent.'') for the ``re-ranking''
oracle. 
Table~\ref{tab:localvsglobal}
lists the performance of our combined models with both local feedback
(Section~\ref{sec:modelloccl}) and global feedback
(Section~\ref{sec:modelgrank}).
The combination
strategy with global rankers uses joint inference and global feedback
(see the description in the previous sub-section).

\begin{table}[t]
\centering
\begin{small}
\begin{tabular}{|r|c|c|c|c||c|c|c|c|}\cline{2-9}
\multicolumn{1}{l|}{} & \multicolumn{4}{|c||}{Combination} & \multicolumn{4}{|c|}{Re-Ranking} \\
\cline{2-9}
\multicolumn{1}{l|}{WSJ} & \#~Args./Sent. & Prec. & Recall & F$_1$ & \#~Frames/Sent. & Prec. & Recall & F$_1$ \\
\hline
~~~~~C2 & 8.53 & 99.34\% & 82.71\% & 90.27 & 3.16 & 88.63\% & 81.77\% & 85.07 \\
\hline
   C4 & 9.78 & 99.47\% & 87.26\% & 92.96 & 4.44 & 91.08\% & 86.12\% & 88.53 \\
\hline
   C6 & 10.23 & 99.47\% & 88.02\% & 93.39 & 7.21 & 92.14\% & 86.57\% & 89.27 \\
\hline
   C8 & 10.74 & 99.48\% & 88.63\% & 93.75 & 8.11 & 92.88\% & 87.33\% & 90.02 \\
\hline
  C10 & 11.33 & {\bf 99.50}\% & {\bf 89.02}\% & {\bf 93.97} & 8.97 & 93.31\% & 87.71\% & 90.42 \\
\hline
%\end{tabular}
%\vspace{0.02in}
%\begin{tabular}{|r|c|c|c|c||c|c|c|c|}
%\cline{2-9}
\multicolumn{9}{l}{Brown}\\% & \#~Args./Sent. & Prec. & Recall & F$_1$ & \#~Frames/Sent. & Prec. & Recall & F$_1$ \\
\hline
~~~C2 & 7.42 & 99.62\% & 71.34\% & 83.14 & 3.02 & 82.45\% & 70.37\% & 75.94 \\
\hline
   C4 & 8.99 & 99.65\% & 77.58\% & 87.24 & 4.55 & 86.01\% & 75.98\% & 80.68 \\
\hline
   C6 & 9.62 & 99.65\% & 79.38\% & 88.37 & 7.09 & 88.19\% & 76.80\% & 82.10 \\
\hline
   C8 & 10.24 & 99.66\% & 80.52\% & 89.08 & 8.19 & 88.95\% & 78.04\% & 83.14 \\
\hline
  C10 & 10.86 & {\bf 99.66}\% & {\bf 81.72}\% & {\bf 89.80} & 9.21 & 89.65\% & 79.19\% & 84.10 \\
\hline
\end{tabular}
\end{small}
\caption{Performance upper limits determined by the oracle systems on
  the 10 best systems at CoNLL-2005. C$k$ stands for combination of
  the top $k$ systems from Table~\ref{tab:10best}. ``\#~Args./Sent.''
  indicates the average number of candidate arguments per sentence for
  the combination oracle; ``\#~Frames/Sent.'' indicates the average
  number of candidate frames per sentence for the re-ranking oracle.
  The latter can be larger than the number of systems in the combination because on average there are multiple predicates per sentence.} 
\label{tab:oracletop10}
\end{table}

\begin{table}[t]
\centering
\begin{small}
\begin{tabular}{|r|c|c|c|c||c|c|c|c|}\cline{2-9}
\multicolumn{1}{l|}{} & \multicolumn{4}{|c||}{Local ranker} & \multicolumn{4}{|c|}{Global ranker}\\
\cline{2-9}
\multicolumn{1}{l|}{WSJ} & PProps & Prec. & Recall & F$_1$ & PProps & Prec. & Recall & F$_1$\\
\hline
~~~~~C2 & 50.69\% & 86.60\% & 73.90\% & 79.75{\scriptsize$\pm 0.7$} & 52.74 & 84.07\% & 75.38\% & 79.49{\scriptsize$\pm 0.7$} \\
\hline
   C4 & 55.14\% & 86.67\% & 76.63\% & 81.38{\scriptsize$\pm 0.7$} & 54.95 & 84.00\% & 77.19\% & 80.45{\scriptsize$\pm 0.7$} \\
\hline
   C6 & 54.85\% & 87.45\% & 76.34\% & {\bf 81.52}{\scriptsize$\pm 0.6$} & {\bf 55.21} & 84.24\% & 77.41\% & 80.68{\scriptsize$\pm 0.7$} \\
\hline
   C8 & 54.36\% & {\bf 87.49}\% & 76.12\% & 81.41{\scriptsize$\pm 0.6$} & 55.00 & 84.42\% & 77.10\% & 80.59{\scriptsize$\pm 0.7$} \\
\hline
  C10 & 53.90\% & 87.48\% & 75.81\% & 81.23{\scriptsize$\pm 0.6$} & 54.76 & 84.02\% & {\bf 77.44}\% & 80.60{\scriptsize$\pm 0.7$} \\
\hline
%\end{tabular}
%\vspace{0.02in}
%\begin{tabular}{|r|c|c|c|c||c|c|c|c|}
%\cline{2-9}
\multicolumn{9}{l}{Brown}\\% & PProps & Prec. & Recall & F$_1$ & PProps & Prec. & Recall & F$_1$\\
\hline
~~~C2 & 32.71\% & 79.56\% & 60.45\% & 68.70{\scriptsize$\pm 1.8$} & 35.32 & 74.88\% & 62.43\% & 68.09{\scriptsize$\pm 2.0$} \\
\hline
   C4 & 35.95\% & 80.27\% & 63.16\% & 70.69{\scriptsize$\pm 2.0$} & {\bf 39.30} & 75.63\% & 64.45\% & 69.59{\scriptsize$\pm 2.2$} \\
\hline
   C6 & 35.32\% & 80.94\% & 62.24\% & 70.37{\scriptsize$\pm 1.8$} & 37.44 &76.12\% & 64.58\% & 69.88{\scriptsize$\pm 2.0$} \\
\hline
   C8 & 35.95\% & 81.98\% & 61.87\% & 70.52{\scriptsize$\pm 2.2$} & 38.43 & 76.40\% & 64.08\% & 69.70{\scriptsize$\pm 2.2$} \\
\hline
  C10 & 36.32\% & {\bf 82.61}\% & 61.97\% & {\bf 70.81}{\scriptsize$\pm 2.0$} & 37.44 & 75.94\% & {\bf 64.68}\% & 69.86{\scriptsize$\pm 2.0$} \\
\hline
\end{tabular}
\end{small}
\caption{Local versus global ranking for combinations of the 10 best
  systems at CoNLL-2005. C$k$ stands for combination of the top $k$
  systems from Table~\ref{tab:10best}.} 
\label{tab:localvsglobal}
\end{table}

We can draw several conclusions from these experiments. 
First, the performance upper limit of re-ranking is always lower than
that that of the argument-based combination strategy, even when the number of candidates
is large. For example, when all 10 individual models are used, the
F$_1$ upper limit of our approach in the Brown corpus 
is 89.80 whereas the F$_1$ upper limit for re-ranking is 84.10. 
However, the enhanced potential of our combination approach does not
imply a significant increase in computational cost:
Table~\ref{tab:oracletop10} shows that the number of candidate
arguments that must be handled by combination approaches is not
that much higher than the number of candidate frames input to the
re-ranking system, especially when the number of individual models is
high. For example, when all 10 individual models are used, the
combination approaches must process around 11 arguments per sentence,
whereas re-ranking approaches must handle approximately 9 frames per
sentence.  
The intuition behind this relatively small difference in computational
cost is that, even though the number of arguments is significantly
larger than the number of frames,  
the difference between the number of {\em unique} candidates for the
two approaches is not high because the probability of repeated
arguments is higher than the probability of repeated frames. 

The second conclusion is that all our
combination models boost the performance of the corresponding 
individual systems.
For example, the best 4-system combination achieves an F$_1$
score approximately 2 points higher than the best individual model in
both the WSJ and Brown corpus. As expected, the combination models
reach a performance plateau around 4-6 individual systems, when the
quality of the individual models starts to drop significantly.
Nevertheless, considering that the top 4 individual systems use
combination strategies themselves and the amount of training data for
this experiment was quite small, these results show the good potential
of the combination models analyzed in this paper.

The third observation is that the relation previously observed between
local and global rankers holds: our combination model with local
rankers has better precision, but the model with global rankers always
has better recall and generally better PProps score.
Overall, the model with local rankers obtains
better F$_1$ scores and scales better as the number of
individual systems increases. We discuss these differences in more
detail in the next sub-section. 

Finally, Table~\ref{tab:oracletop10} indicates that the potential recall of
this experiment (shown in the left-most block in the table) is higher
than the potential recall when combining our three individual SRL
systems (see Table~\ref{tab:upper}): 3.8\% higher in the WSJ test set,
and 7.4\% higher in the Brown test set.
This was expected, considering
that both the number and the quality of the candidate arguments in this last
experiment is higher. However, even after this improvement, the
potential recall of our combination strategies is far from
100\%. Thus, combining the solutions of the $N$ best state-of-the-art
SRL systems still does not have the potential to properly solve the
SRL problem. Future work should focus on recall-boosting strategies,
e.g., using candidate arguments of the individual systems {\em
  before} the individual complete solutions are generated, because in
this step many candidate arguments are eliminated.

\subsection{Discussion}
\label{sec:perfdisc}

The experimental results presented in this section indicate that all
proposed combination strategies are successful: all three combination
models provide statistically significant improvements over the
individual models and the baselines in all setups. An immediate (but
somewhat shallow) comparison of the three combination strategies
investigated indicates that: (a) the best combination strategy for the
SRL problem is a max-margin local meta-learner; (b) the global ranking
approach for the meta-learner is important but it does not have the
same contribution as a max-margin strategy; and (c) the
constraint-satisfaction model performs the worst of all the strategies
tried.

However, in most experiments the differences between the combination
approaches investigated are small. A more reasonable observation is
that each combination strategy has its own advantages and
disadvantages and different approaches are suitable for different
applications and data. We discuss these differences below.

If the argument probabilities of individual systems are available, the
combination model based on constraint satisfaction is an attractive
choice: it is a simple, unsupervised strategy that obtains competitive
performance. Furthermore, the constraint satisfaction model provides
an elegant and customizable framework to tune the balance between
precision and recall (see Section~\ref{sec:perfcs}). With this
framework we currently obtain the highest recall of all combination
models: 3.48\% higher than the best recall obtained by the
meta-learning approaches on the WSJ corpus, and 4.83\% higher than the
meta-learning models on the Brown corpus.
The higher recall implies also higher percentage of predicates that
are completely correctly annotated: the best ``PProps'' numbers in
Table~\ref{tab:localglobal} are the best of all combination strategies.
The cause for this high difference in recall in favor of the
constraint satisfaction approach is that the candidate scoring of the
learning-based inference acts implicitly as a filter:
all candidates whose score
$-$i.e., the classifier confidence that the candidate is part of the 
correct solution$-$ is negative are discarded, which negatively
affects the overall recall. 
Hence, constraint satisfaction is a better
solution for SRL-based NLP applications which require that
predicate-argument frames be extracted with high recall. For example,
in Information Extraction, predicate-argument tuples are filtered with
subsequent high-precision, domain-specific constraints~\cite{sur03},
hence it is paramount that the SRL model have high recall.

Nevertheless, in many cases the argument probabilities of the individual SRL
models are not available, either because the models do not generate
them, e.g., rule-based systems, or because the individual models are
available only as black boxes, which do not offer access to internal
information. Under these conditions, we showed that combination
strategies based on meta-learning are a viable alternative. In fact,
these approaches obtain the highest F$_1$ scores (see
Section~\ref{sec:perfwofeed}) and obtain excellent performance even
with small amounts of training data (see
Section~\ref{sec:perflarger}). As previously mentioned, because
candidate scoring acts as filter, the learning-based inference tends
to favor precision over recall: their precision is 2.82\% higher than
the best precision of the constraint-satisfaction models in the WSJ
corpus, and 3.57\% higher in the Brown corpus.
This preference for precision over
recall is more pronounced in the learning-based inference with local
rankers (Section~\ref{sec:perfwofeed}) than in the inference model
with global rankers (Section~\ref{sec:perfwfeed}). 
Our hypothesis for what causes the global-ranking model to be less
precision-biased is that in this configuration the ratio of errors on
positive versus negative samples is more balanced. Thinking in the
strategy that Perceptron follows, a local approach updates at every
candidate with incorrect prediction sign, whereas a global approach
only updates at candidates that should or should not be in the complete
solution, after enforcing the domain constraints.
In other words, the number of negative updates $-$which drives the
precision bias$-$ is reduced in the global approach, because some of
the false positives generated by the ranking classifiers are
eliminated by the domain constraints.
Thus, because candidate scoring is trained to optimize accuracy, fewer
candidate arguments will be eliminated by the meta-learner with global
rankers, which translates into a better balance between precision and
recall.

\begin{table}[t]
\centering
\begin{tabular}{|r|c|c|c|c|}\cline{2-5}
\cline{2-5}
\multicolumn{1}{l|}{WSJ} & PProps & Prec. & Recall & F$_1$ \\
\hline
global feedback & +3.10\% & -1.96\% & +1.99\% & +0.27 \\
max margin & +0.95\% & +0.67\% & +0.36\% & +0.49 \\
\hline
%\end{tabular}
%\begin{tabular}{|r|c|c|c|c|}\cline{2-5}
%\cline{2-5}
\multicolumn{5}{l}{Brown}\\% & PProps & Prec. & Recall & F$_1$ \\
\hline
global feedback & +2.62\% & -2.71\% & +2.25\% &  +0.38\\
max margin & +1.00\% & +1.13\% & +0.55\% & +0.78 \\
\hline
\end{tabular}

\caption{Contribution of global feedback and max margin to the
  learning-based inference. The baseline is the ``Pred-by-pred,
  local'' model in Table~\ref{tab:perfwfeed}.} 
\label{tab:mmvsgf}
\end{table}

Another important conclusion of our analysis of global versus local
ranking for the learning-based inference is that a max-margin approach
for the candidate scoring classifiers is more important than having
global feedback for inference. In fact, considering that the only
difference between the model with predicate-by-predicate inference
with local feedback in Section~\ref{sec:perfwfeed} (``Pred-by-pred,
local'') versus the best model in Section~\ref{sec:perfwofeed} (+FS6)
is that the latter uses SVM classifiers whereas the former uses
Perceptron, we can compute the exact contribution of both max margin
and global feedback\footnote{The contribution of global feedback is
  given by the model with joint inference and global feedback (``Full sentence,
  global'') in Section~\ref{sec:perfwfeed}.}. For convenience, we summarize
this analysis in Table~\ref{tab:mmvsgf}. That table
indicates that max margin yields a consistent improvement of both
precision and recall, whereas the contribution of global feedback is
more in reducing the difference between precision and recall by
boosting recall and decreasing precision. The benefit of max-margin
classifiers is even more evident in Table~\ref{tab:localvsglobal},
which shows that the local-ranking model with max-margin classifiers
generalizes better than the global-ranking model when 
the amount of training data is reduced significantly.

Even though in this paper we have analyzed several combination
approaches with three independent implementations, the proposed models
are in fact compatible with each other. Various combinations of the
proposed strategies are immediately possible. For example, the
constraint satisfaction model can be applied on the output probabilities of
the candidate scoring component introduced in
Section~\ref{sec:modelwofeed}. Such a model eliminates the dependency
on the output scores of the individual SRL models but retains all
the advantages of the constraint satisfaction model, e.g., the formal
framework to tune the balance between precision and recall. 
Another possible combination of the approaches introduced in this
paper is to use max-margin classifiers in the learning-based inference
with global feedback, e.g., by using a global training method for
margin maximization such as SVMstruct \cite{tsochan04}.
This model would indeed have an increased training
time\footnote{This was the main reason why we chose Perceptron for the
  proposed online strategies.}, but could leverage the advantages of
both max-margin classifiers and inference with global feedback
(summarized in Table~\ref{tab:mmvsgf}). 
Finally, another attractive approach is stacking, i.e., $N$ levels of
chained meta-learning. For example, we could cascade the
learning-based inference model with global rankers, which boosts
recall, with the learning-based inference with local rankers, which
favors precision.

\section{Related Work}
\label{sec:related}

The 4 best performing systems at the CoNLL-2005 shared task included a
combination of different base subsystems to increase robustness and to
gain coverage and independence from parse errors. Therefore, they are
closely related to the work of this paper. The first four rows in
Table \ref{tab:10best} summarize their results under exactly the same
experimental setting as the one used in this paper. 

\citeA{koo05} used a 2 layer architecture close to ours. The pool
of candidates is generated by: (a) running a full syntax SRL system on
alternative input information (Collins parsing, and $5$-best trees
from Charniak's parser),
and (b): taking all candidates that pass a filter from the set of
different parse trees.
The combination of candidates is performed in
an elegant global inference procedure as constraint satisfaction,
which, formulated as Integer Linear Programming, is solved
efficiently. 
This is different from our work, where we ``break'' complete solutions
from any number of SRL systems and we also investigate a meta-learning
combination approach in addition to the ILP inference. 
Koomen et al.'s system was the best performing system at CoNLL-2005
(see Table~\ref{tab:10best}). 

\citeA{hag05} implemented double re-ranking on top of several outputs
from a base SRL model.
% to select the most probable solution among a set of candidates. 
The re-ranking is performed, first, on a set of $n$-best
solutions obtained by the base system run on a single parse tree, and,
then, on the set of best-candidates coming from the $n$-best parse
trees. This was the second-best system at CoNLL-2005 (third row in
Table~\ref{tab:10best}).
Compared to our decomposition and re-combination approach, the
re-ranking setting has the advantage of allowing the definition of
global features that apply to complete candidate solutions. According
to a follow-up work by the same authors \cite{toutanova05}, these
global features are the source of the major performance improvements
of the re-ranking system. In contrast, we focus more on features that
exploit the redundancy between the individual models, e.g., overlap
between individual candidate arguments, and we add global information
at frame level only from the complete solutions provided by individual
models.
The main drawback of re-ranking compared to our approach is that the
different individual solutions can not be combined because re-ranking
is forced to select a complete candidate solution. This implies that
its overall performance strongly depends on the ability of the base
model to generate the complete correct solution in the set of $n$-best
candidates. This drawback is evident in the lower performance upper
limit of the re-ranking approach (see Tables~\ref{tab:upper}
and~\ref{tab:oracletop10}) and in the performance of the actual system
---our best combination strategy achieves an F$_1$ score over 2 points
higher than Haghighi et al. in both WSJ and Brown\footnote{Recently,
\citeA{yih06} reported improved numbers for this system: 80.32 F$_1$
for WSJ and 68.81 for Brown. However, these numbers are not directly
comparable with the systems presented in this paper because they
fixed a significant bug in the representation of quotes in the input
data, a bug that is still present in our data.}.

Finally, \citeA{pra05b} followed a stacking approach by learning two
individual systems based on full syntax, whose outputs are used to
generate features to feed the training stage of a final chunk-by-chunk
SRL system. Although the fine granularity of the chunking-based system
allows to recover from parsing errors, we find this combination scheme
quite ad-hoc because it forces to break argument candidates
into chunks in the last stage.

Outside of the CoNLL shared task evaluation, \citeA{rot05} reached the
conclusion that the quality of the local argument classifiers is more
important than the global feedback from the inference component. 
This is also one
of the conclusions drawn by this paper. Our contribution is that we have
shown that this hypothesis holds in a more complex framework: combination
of several state-of-the-art individual models, whereas Roth and Yih
experimented with a single individual model, only numbered arguments, and
a slightly simplified problem representation: B-I-O over basic chunks.
Additionally, our more detailed experiments allowed us to show clearly
that the contribution of max margin is higher than that of global learning
in several corpora and for several combinations of individual systems.

\citeA{pun05} showed that the performance of individual SRL models
(particularly argument identification) is significantly improved when
full parsing is used and argument boundaries are restricted to match
syntactic constituents (similarly to our Model~3). We believe that the
approach used by our Models 1 and 2, where candidate arguments do not
have to match a single syntactic constituent, has increased robustness
because it has a built-in mechanism to handle some syntax errors, when
an argument constituent is incorrectly fragmented into multiple
phrases. Our empirical results support this claim: Model~2 performs
better than both Model~3 and the models proposed by Punyakanok et al. A
second advantage of the strategy proposed in this paper is that the
same model can be deployed using full syntax (Model~2) or partial
syntax (Model~1). 

\citeA{pra05c} implement a SRL combination strategy at constituent
level that, similarly to our approach, combines different syntactic
views of the data based on full and partial syntactic
analysis. However, unlike our approach, Pradhan et al.'s work uses
only a simple greedy inference strategy based on the probabilities of
the candidate arguments, whereas in this paper we introduce and
analyze three different combination algorithms. Our analysis yielded a
combination system that outperforms the current state of the art. 

Previous work in the more general field of predicting structures in
natural language texts has indicated that the combination of several
individual models improves overall performance in the given task.
\citeA{col00} first proposed a learning layer based on
ranking to improve the performance of a generative syntactic parser.
In that approach, a reranker was trained to select the best solution
from a pool of solutions produced by the generative
parser. In doing so, the reranker dealt with complete parse trees, and
represented them with rich features that exploited dependencies not
considered in the generative method. On the other hand, it was
computationally feasible to train the reranker, because the base
method reduced the number of possible parse trees for a sentence from
an exponential number (w.r.t. sentence length) to a few tens.
More recently, global discriminative learning methods for predicting
structures have been proposed
\cite{lafferty01,col02c,col04,taskar03,taskar04,tsochan04}. All of
them train a single discriminative ranking function to detect
structures in a sentence. A major property of these methods is that
they model the problem discriminatively, so that arbitrary and rich
representations of structures can be used. 
Furthermore, the training process in these methods is global, in that
parameters are set to maximize measures not only related to local
accuracies (i.e., on recognizing parts of a structure), but also related
to the global accuracy (i.e., on recognizing complete structures).
In this article, the use of global and rich representations is also a
major motivation. 

\vspace{-0.1in}
\section{Conclusions}
\label{sec:conclusions}

This paper introduces and analyzes three combination strategies in the
context of semantic role labeling: the first model implements an inference
strategy with constraint satisfaction using integer linear programming,
the second uses inference based on learning where the candidates are
scored using discriminative classifiers using only local information, and
the third and last inference model builds on the previous strategy by
adding global feedback from the conflict resolution component to the
ranking classifiers. The meta-learners used by the inference process are
developed with a rich set of features that includes voting statistics
$-$i.e., how many individual systems proposed a candidate argument$-$
overlap with arguments from the same and other predicates in the sentence,
structure and distance information coded using partial and full syntax,
and probabilities from the individual SRL models (if available). To our
knowledge, this is the first work that: (a) introduces a thorough
inference model based on learning for semantic role labeling, and (b)
performs a comparative analysis of several inference strategies in
the context of SRL.

The results presented suggest that the strategy of decomposing
individual solutions and performing a learning-based re-combination
for constructing the final solution has advantages over other
approaches, e.g., re-ranking a set of complete candidate solutions.
Of course, this is a task-dependant conclusion. In the case of
semantic role labeling, our approach is relatively simple since the
re-combination of argument candidates has to fulfill only a few set of
structural constraints to generate a consistent solution. If the
target structure is more complex (e.g., a full parse tree) the
re-combination step might be too complex from both the
learning and search perspectives.

Our evaluation indicates that all proposed combination approaches are
successful: they all provide significant improvements
over the best individual model and several baseline combination algorithms
in all setups. Out of the three combination strategies investigated,
the best F$_1$ score is obtained by the learning-based
inference using max-margin classifiers. 
While all the proposed approaches have their own advantages and drawbacks
(see Section~\ref{sec:perfdisc} for a detailed discussion of
differences among the proposed inference models) several 
important features of a state-of-the-art SRL combination strategy emerge 
from this analysis:
(i) individual models should be combined at the granularity of candidate 
arguments rather than at the granularity of complete solutions or frames; 
(ii)  the best combination strategy uses an inference model based in learning;
(iii) the learning-based inference benefits from max-margin classifiers and global feedback,
and (iv) the inference at sentence level (i.e., considering all predicates at
the same time) proves only slightly useful when the learning is
performed also globally, using feedback from the complete solution
after inference.

Last but not least, the results obtained with the best combination strategy
developed in this work outperform the current state of the art. These results
are empirical proof that a SRL system with good performance can be built by
combining a small number (three in our experiments) of relatively simple
SRL models.

\acks{We would like to thank the JAIR reviewers for their valuable comments.\\
This research has been partially supported by the European
  Commission (CHIL project, IP-506909; PASCAL Network,
  IST-2002-506778) and the Spanish Ministry of Education and Science
  (TRANGRAM, TIN2004-07925-C03-02). Mihai Surdeanu is a research
  fellow within the Ram{\'o}n y Cajal program of the Spanish Ministry
  of Education and Science. We are also grateful to Dash Optimization for the free academic use of Xpress-MP.

}

\vskip 0.2in
\bibliography{surdeanu07a}

\begin{thebibliography}{}

\bibitem[\protect\BCAY{Bishop}{Bishop}{1995}]{bish95}
Bishop, C. \BBOP1995\BBCP.
\newblock {\Bem Neural Networks for Pattern Recognition}.
\newblock Oxford University Press.

\bibitem[\protect\BCAY{Boas}{Boas}{2002}]{boa02}
Boas, H.~C. \BBOP2002\BBCP.
\newblock \BBOQ Bilingual framenet dictionaries for machine translation\BBCQ\
\newblock In {\Bem Proceedings of LREC 2002}.

\bibitem[\protect\BCAY{Carreras\ \BBA\ M\`{a}rquez}{Carreras\ \BBA\
  M\`{a}rquez}{2004}]{car04}
Carreras, X.\BBACOMMA\  \BBA\ M\`{a}rquez, L. \BBOP2004\BBCP.
\newblock \BBOQ Introduction to the {CoNLL-2004} shared task: Semantic role
  labeling\BBCQ\
\newblock In {\Bem Proceedings of CoNLL 2004}.

\bibitem[\protect\BCAY{Carreras\ \BBA\ M{\`a}rquez}{Carreras\ \BBA\
  M{\`a}rquez}{2005}]{car05}
Carreras, X.\BBACOMMA\  \BBA\ M{\`a}rquez, L. \BBOP2005\BBCP.
\newblock \BBOQ Introduction to the conll-2005 shared task: Semantic role
  labeling\BBCQ\
\newblock In {\Bem Proceedings of CoNLL-2005}.

\bibitem[\protect\BCAY{Carreras, M\`{a}rquez,\ \BBA\ Chrupa{\l}a}{Carreras
  et~al.}{2004}]{car04b}
Carreras, X., M\`{a}rquez, L., \BBA\ Chrupa{\l}a, G. \BBOP2004\BBCP.
\newblock \BBOQ Hierarchical recognition of propositional arguments with
  perceptrons\BBCQ\
\newblock In {\Bem Proceedings of CoNLL 2004 Shared Task}.

\bibitem[\protect\BCAY{Charniak}{Charniak}{2000}]{cha00}
Charniak, E. \BBOP2000\BBCP.
\newblock \BBOQ A maximum-entropy-inspired parser\BBCQ\
\newblock In {\Bem Proceedings of NAACL}.

\bibitem[\protect\BCAY{Collins}{Collins}{1999}]{col99b}
Collins, M. \BBOP1999\BBCP.
\newblock {\Bem Head-Driven Statistical Models for Natural Language Parsing}.
\newblock PhD Dissertation, University of Pennsylvania.

\bibitem[\protect\BCAY{Collins}{Collins}{2000}]{col00}
Collins, M. \BBOP2000\BBCP.
\newblock \BBOQ Discriminative reranking for natural language parsing\BBCQ\
\newblock In {\Bem Proceedings of the 17th International Conference on Machine
  Learning, ICML-00}, Stanford, CA USA.

\bibitem[\protect\BCAY{Collins}{Collins}{2002}]{col02c}
Collins, M. \BBOP2002\BBCP.
\newblock \BBOQ Discriminative training methods for hidden markov models:
  Theory and experiments with perceptron algorithms\BBCQ\
\newblock In {\Bem Proceedings of the SIGDAT Conference on Empirical Methods in
  Natural Language Processing, EMNLP-02}.

\bibitem[\protect\BCAY{Collins}{Collins}{2004}]{col04}
Collins, M. \BBOP2004\BBCP.
\newblock \BBOQ Parameter estimation for statistical parsing models: Theory and
  practice of distribution-free methods\BBCQ\
\newblock In Bunt, H., Carroll, J., \BBA\ Satta, G.\BEDS, {\Bem New
  Developments in Parsing Technology}, \BCH~2. Kluwer.

\bibitem[\protect\BCAY{Collins\ \BBA\ Duffy}{Collins\ \BBA\
  Duffy}{2002}]{col02a}
Collins, M.\BBACOMMA\  \BBA\ Duffy, N. \BBOP2002\BBCP.
\newblock \BBOQ New ranking algorithms for parsing and tagging: Kernels over
  discrete structures, and the voted perceptron\BBCQ\
\newblock In {\Bem {Proceedings of the 40th Annual Meeting of the Association
  for Computational Linguistics, ACL'02}}.

\bibitem[\protect\BCAY{Crammer\ \BBA\ Singer}{Crammer\ \BBA\
  Singer}{2003a}]{cramSi03b}
Crammer, K.\BBACOMMA\  \BBA\ Singer, Y. \BBOP2003a\BBCP.
\newblock \BBOQ A family of additive online algorithms for category
  ranking\BBCQ\
\newblock {\Bem Journal of Machine Learning Research}, {\Bem 3}, 1025--1058.

\bibitem[\protect\BCAY{Crammer\ \BBA\ Singer}{Crammer\ \BBA\
  Singer}{2003b}]{cramSi03a}
Crammer, K.\BBACOMMA\  \BBA\ Singer, Y. \BBOP2003b\BBCP.
\newblock \BBOQ Ultraconservative online algorithms for multiclass
  problems\BBCQ\
\newblock {\Bem Journal of Machine Learning Research}, {\Bem 3}, 951--991.

\bibitem[\protect\BCAY{Freund\ \BBA\ Schapire}{Freund\ \BBA\
  Schapire}{1999}]{frSch99}
Freund, Y.\BBACOMMA\  \BBA\ Schapire, R.~E. \BBOP1999\BBCP.
\newblock \BBOQ Large margin classification using the perceptron
  algorithm\BBCQ\
\newblock {\Bem Machine Learning}, {\Bem 37\/}(3), 277--296.

\bibitem[\protect\BCAY{Gildea\ \BBA\ Jurafsky}{Gildea\ \BBA\
  Jurafsky}{2002}]{gil02}
Gildea, D.\BBACOMMA\  \BBA\ Jurafsky, D. \BBOP2002\BBCP.
\newblock \BBOQ Automatic labeling of semantic roles\BBCQ\
\newblock {\Bem Computational Linguistics}, {\Bem 28\/}(3).

\bibitem[\protect\BCAY{Gildea\ \BBA\ Palmer}{Gildea\ \BBA\
  Palmer}{2002}]{gil02b}
Gildea, D.\BBACOMMA\  \BBA\ Palmer, M. \BBOP2002\BBCP.
\newblock \BBOQ The necessity of syntactic parsing for predicate argument
  recognition\BBCQ\
\newblock In {\Bem Proceedings of the 40th Annual Conference of the Association
  for Computational Linguistics (ACL-02)}.

\bibitem[\protect\BCAY{Hacioglu, Pradhan, Ward, Martin,\ \BBA\
  Jurafsky}{Hacioglu et~al.}{2004}]{hacioglu04}
Hacioglu, K., Pradhan, S., Ward, W., Martin, J.~H., \BBA\ Jurafsky, D.
  \BBOP2004\BBCP.
\newblock \BBOQ Semantic role labeling by tagging syntactic chunks\BBCQ\
\newblock In {\Bem Proceedings of the 8th Conference on Computational Natural
  Language Learning (CoNLL-2004)}.

\bibitem[\protect\BCAY{Haghighi, Toutanova,\ \BBA\ Manning}{Haghighi
  et~al.}{2005}]{hag05}
Haghighi, A., Toutanova, K., \BBA\ Manning, C. \BBOP2005\BBCP.
\newblock \BBOQ A joint model for semantic role labeling\BBCQ\
\newblock In {\Bem Proceedings of CoNLL-2005 Shared Task}.

\bibitem[\protect\BCAY{Koomen, Punyakanok, Roth,\ \BBA\ Yih}{Koomen
  et~al.}{2005}]{koo05}
Koomen, P., Punyakanok, V., Roth, D., \BBA\ Yih, W. \BBOP2005\BBCP.
\newblock \BBOQ Generalized inference with multiple semantic role labeling
  systems\BBCQ\
\newblock In {\Bem Proceedings of CoNLL-2005 Shared Task}.

\bibitem[\protect\BCAY{Lafferty, McCallum,\ \BBA\ Pereira}{Lafferty
  et~al.}{2001}]{lafferty01}
Lafferty, J., McCallum, A., \BBA\ Pereira, F. \BBOP2001\BBCP.
\newblock \BBOQ Conditonal random fields: Probabilistic models for segmenting
  and labeling sequence data\BBCQ\
\newblock In {\Bem {Proceedings of the 18th International Conference on Machine
  Learning, ICML-01}}.

\bibitem[\protect\BCAY{Marcus, Santorini,\ \BBA\ Marcinkiewicz}{Marcus
  et~al.}{1994}]{mar94}
Marcus, M., Santorini, B., \BBA\ Marcinkiewicz, M. \BBOP1994\BBCP.
\newblock \BBOQ Building a large annotated corpus of {English}: The {Penn
  Treebank}\BBCQ\
\newblock {\Bem Computational Linguistics}, {\Bem 19\/}(2).

\bibitem[\protect\BCAY{M{\`a}rquez, Comas, Gim{\'e}nez,\ \BBA\
  Catal{\`a}}{M{\`a}rquez et~al.}{2005}]{mar05}
M{\`a}rquez, L., Comas, P., Gim{\'e}nez, J., \BBA\ Catal{\`a}, N.
  \BBOP2005\BBCP.
\newblock \BBOQ Semantic role labeling as sequential tagging\BBCQ\
\newblock In {\Bem Proceedings of CoNLL-2005 Shared Task}.

\bibitem[\protect\BCAY{Melli, Wang, Liu, Kashani, Shi, Gu, Sarkar,\ \BBA\
  Popowich}{Melli et~al.}{2005}]{mwl+05}
Melli, G., Wang, Y., Liu, Y., Kashani, M.~M., Shi, Z., Gu, B., Sarkar, A.,
  \BBA\ Popowich, F. \BBOP2005\BBCP.
\newblock \BBOQ Description of {SQUASH}, the {SFU} question answering summary
  handler for the {DUC-2005} summarization task\BBCQ\
\newblock In {\Bem Proceedings of Document Understanding Workshop, HLT/EMNLP
  Annual Meeting}.

\bibitem[\protect\BCAY{Narayanan\ \BBA\ Harabagiu}{Narayanan\ \BBA\
  Harabagiu}{2004}]{nar04}
Narayanan, S.\BBACOMMA\  \BBA\ Harabagiu, S. \BBOP2004\BBCP.
\newblock \BBOQ Question answering based on semantic structures\BBCQ\
\newblock In {\Bem International Conference on Computational Linguistics
  (COLING 2004)}.

\bibitem[\protect\BCAY{Noreen}{Noreen}{1989}]{nor89}
Noreen, E.~W. \BBOP1989\BBCP.
\newblock {\Bem Computer-Intensive Methods for Testing Hypotheses}.
\newblock John Wiley \& Sons.

\bibitem[\protect\BCAY{Palmer, Gildea,\ \BBA\ Kingsbury}{Palmer
  et~al.}{2005}]{pal05}
Palmer, M., Gildea, D., \BBA\ Kingsbury, P. \BBOP2005\BBCP.
\newblock \BBOQ The {Proposition Bank}: An annotated corpus of semantic
  roles\BBCQ\
\newblock {\Bem Computational Linguistics}, {\Bem 31\/}(1).

\bibitem[\protect\BCAY{Ponzetto\ \BBA\ Strube}{Ponzetto\ \BBA\
  Strube}{2006a}]{pon06b}
Ponzetto, S.~P.\BBACOMMA\  \BBA\ Strube, M. \BBOP2006a\BBCP.
\newblock \BBOQ Exploiting semantic role labeling, wordnet and wikipedia for
  coreference resolution\BBCQ\
\newblock In {\Bem Proceedings of the Human Language Technolgy Conference of
  the North American Chapter of the Association for Computational Linguistics}.

\bibitem[\protect\BCAY{Ponzetto\ \BBA\ Strube}{Ponzetto\ \BBA\
  Strube}{2006b}]{pon06a}
Ponzetto, S.~P.\BBACOMMA\  \BBA\ Strube, M. \BBOP2006b\BBCP.
\newblock \BBOQ Semantic role labeling for coreference resolution\BBCQ\
\newblock In {\Bem Companion Volume of the Proceedings of the 11th Meeting of
  the European Chapter of the Association for Computational Linguistics}.

\bibitem[\protect\BCAY{Pradhan, Hacioglu, Krugler, Ward, Martin,\ \BBA\
  Jurafsky}{Pradhan et~al.}{2005a}]{pra05}
Pradhan, S., Hacioglu, K., Krugler, V., Ward, W., Martin, J.~H., \BBA\
  Jurafsky, D. \BBOP2005a\BBCP.
\newblock \BBOQ Support vector learning for semantic argument
  classification\BBCQ\
\newblock {\Bem Machine Learning}, {\Bem 60}, 11--39.

\bibitem[\protect\BCAY{Pradhan, Hacioglu, Ward, Martin,\ \BBA\
  Jurafsky}{Pradhan et~al.}{2005b}]{pra05b}
Pradhan, S., Hacioglu, K., Ward, W., Martin, J.~H., \BBA\ Jurafsky, D.
  \BBOP2005b\BBCP.
\newblock \BBOQ Semantic role chunking combining complementary syntactic
  views\BBCQ\
\newblock In {\Bem Proceedings of CoNLL-2005}.

\bibitem[\protect\BCAY{Pradhan, Ward, Hacioglu, Martin,\ \BBA\
  Jurafsky}{Pradhan et~al.}{2005c}]{pra05c}
Pradhan, S., Ward, W., Hacioglu, K., Martin, J.~H., \BBA\ Jurafsky, D.
  \BBOP2005c\BBCP.
\newblock \BBOQ Semantic role labeling using different syntactic views\BBCQ\
\newblock In {\Bem Proceedings of the 43rd Annual Conference of the Association
  for Computational Linguistics}.

\bibitem[\protect\BCAY{Punyakanok, Roth,\ \BBA\ Yih}{Punyakanok
  et~al.}{2005}]{pun05}
Punyakanok, V., Roth, D., \BBA\ Yih, W. \BBOP2005\BBCP.
\newblock \BBOQ The necessity of syntactic parsing for semantic role
  labeling\BBCQ\
\newblock In {\Bem Proceedings of the International Joint Conference on
  Artificial Intelligence (IJCAI)}.

\bibitem[\protect\BCAY{Punyakanok, Roth, Yih,\ \BBA\ Zimak}{Punyakanok
  et~al.}{2004}]{roth04b}
Punyakanok, V., Roth, D., Yih, W., \BBA\ Zimak, D. \BBOP2004\BBCP.
\newblock \BBOQ Semantic role labeling via integer linear programming
  inference\BBCQ\
\newblock In {\Bem Proceedings of the International Conference on Computational
  Linguistics (COLING'04)}.

\bibitem[\protect\BCAY{Rosenblatt}{Rosenblatt}{1958}]{rosen58}
Rosenblatt, F. \BBOP1958\BBCP.
\newblock \BBOQ The perceptron: A probabilistic model for information storage
  and organization in the brain\BBCQ\
\newblock {\Bem Psychological Review}, {\Bem 65}, 386--407.

\bibitem[\protect\BCAY{Roth\ \BBA\ Yih}{Roth\ \BBA\ Yih}{2004}]{roth04a}
Roth, D.\BBACOMMA\  \BBA\ Yih, W. \BBOP2004\BBCP.
\newblock \BBOQ A linear programming formulation for global inference in
  natural language tasks\BBCQ\
\newblock In {\Bem Proceedings of the Annual Conference on Computational
  Natural Language Learning (CoNLL-2004)}, \BPGS\ 1--8, Boston, MA.

\bibitem[\protect\BCAY{Roth\ \BBA\ Yih}{Roth\ \BBA\ Yih}{2005}]{rot05}
Roth, D.\BBACOMMA\  \BBA\ Yih, W. \BBOP2005\BBCP.
\newblock \BBOQ Integer linear programming inference for conditional random
  fields\BBCQ\
\newblock In {\Bem Proceedings of the International Conference on Machine
  Learning (ICML)}.

\bibitem[\protect\BCAY{Schapire\ \BBA\ Singer}{Schapire\ \BBA\
  Singer}{1999}]{sch99}
Schapire, R.~E.\BBACOMMA\  \BBA\ Singer, Y. \BBOP1999\BBCP.
\newblock \BBOQ Improved boosting algorithms using confidence-rated
  predictions\BBCQ\
\newblock {\Bem Machine Learning}, {\Bem 37\/}(3).

\bibitem[\protect\BCAY{Surdeanu, Harabagiu, Williams,\ \BBA\ Aarseth}{Surdeanu
  et~al.}{2003}]{sur03}
Surdeanu, M., Harabagiu, S., Williams, J., \BBA\ Aarseth, P. \BBOP2003\BBCP.
\newblock \BBOQ Using predicate-argument structures for information
  extraction\BBCQ\
\newblock In {\Bem Proceedings of the 41st Annual Meeting of the Association
  for Computational Linguistics (ACL 2003)}.

\bibitem[\protect\BCAY{Taskar, Guestrin,\ \BBA\ Koller}{Taskar
  et~al.}{2003}]{taskar03}
Taskar, B., Guestrin, C., \BBA\ Koller, D. \BBOP2003\BBCP.
\newblock \BBOQ {Max-Margin Markov Networks}\BBCQ\
\newblock In {\Bem {Proceedings of the 17th Annual Conference on Neural
  Information Processing Systems, NIPS-03}}, {Vancouver, Canada}.

\bibitem[\protect\BCAY{Taskar, Klein, Collins, Koller,\ \BBA\ Manning}{Taskar
  et~al.}{2004}]{taskar04}
Taskar, B., Klein, D., Collins, M., Koller, D., \BBA\ Manning, C.
  \BBOP2004\BBCP.
\newblock \BBOQ Max-margin parsing\BBCQ\
\newblock In {\Bem Proceedings of the EMNLP-2004}.

\bibitem[\protect\BCAY{Toutanova, Haghighi,\ \BBA\ Manning}{Toutanova
  et~al.}{2005}]{toutanova05}
Toutanova, K., Haghighi, A., \BBA\ Manning, C. \BBOP2005\BBCP.
\newblock \BBOQ Joint learning improves semantic role labeling\BBCQ\
\newblock In {\Bem Proceedings of the 43rd Annual Meeting of the Association
  for Computational Linguistics (ACL'05)}, \BPGS\ 589--596, Ann Arbor, MI, USA.
  Association for Computational Linguistics.

\bibitem[\protect\BCAY{Tsochantaridis, Hofmann, Joachims,\ \BBA\
  Altun}{Tsochantaridis et~al.}{2004}]{tsochan04}
Tsochantaridis, I., Hofmann, T., Joachims, T., \BBA\ Altun, Y. \BBOP2004\BBCP.
\newblock \BBOQ Support vector machine learning for interdependent and
  structured output spaces\BBCQ\
\newblock In {\Bem Proceedings of the 21st International Conference on Machine
  Learning, ICML-04}.

\bibitem[\protect\BCAY{Xue\ \BBA\ Palmer}{Xue\ \BBA\ Palmer}{2004}]{xue04}
Xue, N.\BBACOMMA\  \BBA\ Palmer, M. \BBOP2004\BBCP.
\newblock \BBOQ Calibrating features for semantic role labeling\BBCQ\
\newblock In {\Bem Proceedings of EMNLP-2004}.

\bibitem[\protect\BCAY{Yih\ \BBA\ Toutanova}{Yih\ \BBA\
  Toutanova}{2006}]{yih06}
Yih, S.~W.\BBACOMMA\  \BBA\ Toutanova, K. \BBOP2006\BBCP.
\newblock \BBOQ Automatic semantic role labeling\BBCQ\
\newblock In {\Bem Tutorial of the Human Language Technolgy Conference of the
  North American Chapter of the Association for Computational Linguistics}.

\bibitem[\protect\BCAY{Younger}{Younger}{1967}]{you67}
Younger, D.~H. \BBOP1967\BBCP.
\newblock \BBOQ Recognition and parsing of context-free languages in $n^3$
  time\BBCQ\
\newblock {\Bem Information and Control}, {\Bem 10\/}(2), 189--208.

\end{thebibliography}
\bibliographystyle{theapa}

\end{document}